\documentclass[letterpaper, 10pt, conference]{ieeeconf}
\IEEEoverridecommandlockouts
\usepackage{amsmath,amssymb,amsfonts}
\usepackage{graphicx}
\usepackage{booktabs}
\usepackage{multirow}
\usepackage{xcolor}
\usepackage[caption=false,font=footnotesize]{subfig}
\usepackage{cite}
\usepackage{url}
\usepackage{tikz}
\usetikzlibrary{arrows.meta,positioning,fit,calc}

\newcommand{\sys}{DreamLedger}

\begin{document}

\title{\LARGE \bf \sys{}: Where to Refuse World-Model Imagination\\Using Execution-Settled Credit}

\author{Xianyao Li$^{1}$, Ruitong Tian$^{1}$, Rui Min$^{2}$, Fang Xu$^{1}$, Eric Jing Du$^{1,\dagger}$%
\thanks{$^{1}$X. Li, R. Tian, F. Xu, and E. J. Du are with the Department of Civil and Coastal Engineering, University of Florida, Gainesville, FL 32611, USA.}%
\thanks{$^{2}$R. Min is with the Department of Mechanical and Aerospace Engineering, University of Florida, Gainesville, FL 32611, USA.}%
\thanks{$^{\dagger}$Corresponding author: Eric Jing Du, \texttt{eric.du@essie.ufl.edu}. First author: \texttt{xianyao.li@ufl.edu}.}}

\maketitle
\thispagestyle{empty}
\pagestyle{empty}

\begin{abstract}
World-model predictions inform robot actions, yet instantaneous reliability signals do not retain the outcomes of comparable past predictions. \sys{} registers consumed predictions as claims, settles them against execution outcomes, and uses persistent execution history from comparable operating conditions, regions, and prediction horizons to estimate credit before future reliance. Replayable records connect each decision to its supporting evidence and eventual outcome. In ten-seed navigation comparisons at matched refusal volume, removing history features or resetting history increases burn rate, measured as failures per consumed prediction. An independent ten-seed manipulation replication at matched refusal volume finds that, relative to random refusal, \sys{} lowers burn rate by 4.8 percentage points (95\% CI: 0.8--8.7) and uses fewer probes. Randomized audits directly measure higher failure rates among denied candidates, and post-warmup shifts isolate the contribution of newly accumulated settlements. Franka experiments establish online deployment through replay of all 1{,}062 prediction uses and demonstrate a prospective gate transition: new failures lower previously high credit below a frozen threshold, triggering refusal before the next action. Task completion and verification cost characterize the trade-offs of these interventions.
\end{abstract}

\section{Introduction}

Trust in a forecaster depends not only on today's forecast but also on a record of forecasts made under comparable conditions. Meteorology has maintained such records for decades~\cite{brier1950}, including condition-dependent verification~\cite{murphy1995}. World models now play an analogous role in robotic decision loops: they predict future states or observations beyond the robot's sensing horizon, and planners may act directly on these predictions. However, planners typically consume them without a persistent record of where, under which conditions, and over what horizons they have been reliable. This paper develops such a record for world-model imagination.

\begin{figure}[t]
\centering
\includegraphics[width=\linewidth]{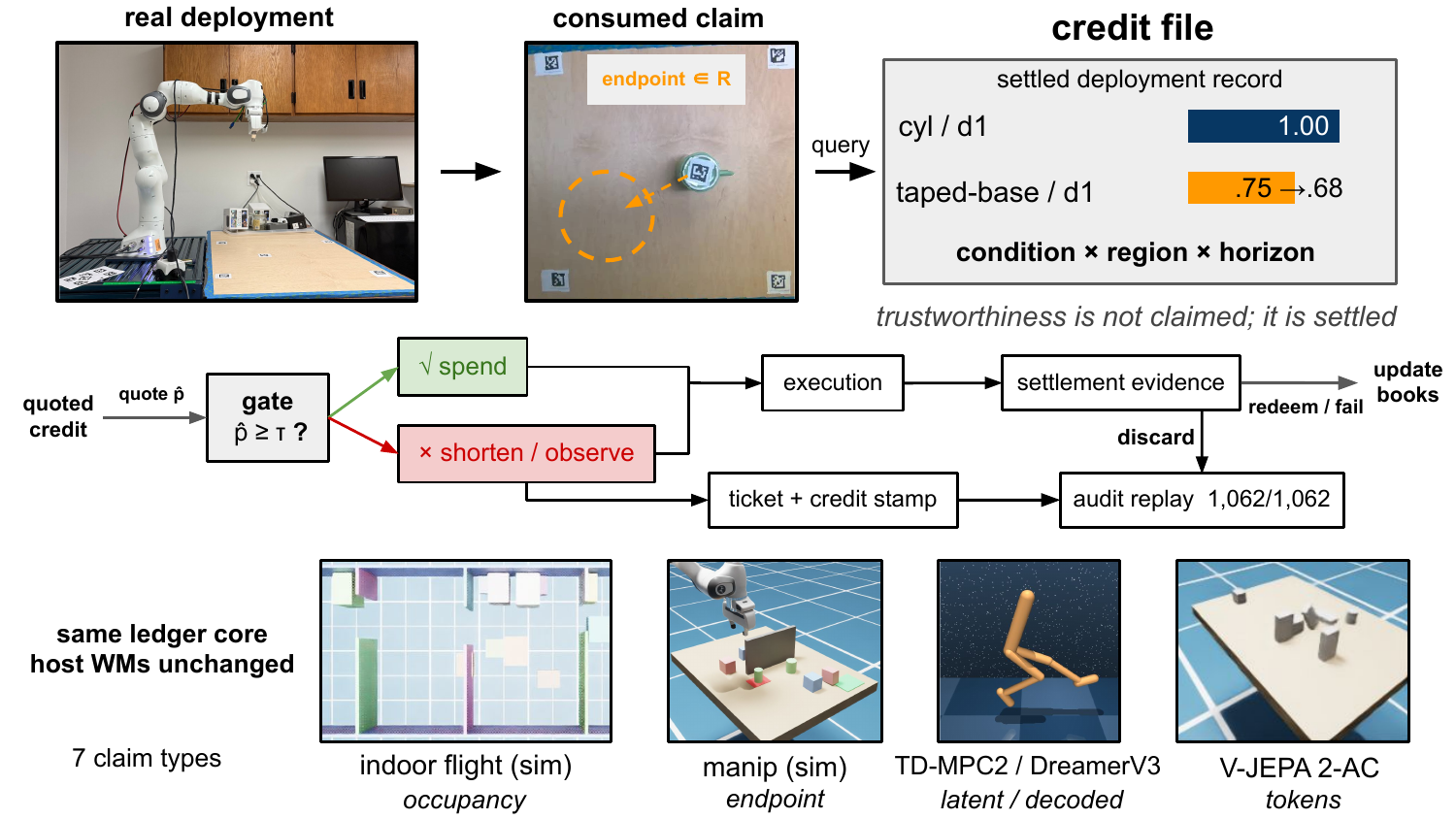}
\caption{Overview of \sys{}. \emph{Top:} In a real Franka tabletop deployment, a consumed prediction is frozen as a \emph{claim} (endpoint within $R$; actual wrist-camera view). Settled history provides $\hat p$ and gates reliance, attributable execution evidence updates contextual history, and every prediction use is audit-replayable (all values are deployment records). \emph{Bottom:} With host models unchanged, the same ledger core spans three simulated domains and three established open-source world models, settling seven claim types (Sec.~\ref{sec:mounts}).}
\label{fig:teaser}
\end{figure}

The need for such a record is especially acute during deployment. Prediction errors can compound over the rollout horizon, and model reliability can degrade under distribution shift~\cite{zidan2026survey}. Model-internal confidence may also fail to reflect the task risk induced by an erroneous prediction~\cite{wang2026feedback}. In recurrent world models, ensemble disagreement can miss compounding rollout error~\cite{berger2026biased}, allowing confident but incorrect imagined futures to create a false sense of safety~\cite{liu2026security}. Regulatory frameworks increasingly emphasize event logging for high-risk AI systems (e.g., EU AI Act, Art.~12~\cite{euaiact2024}), yet conventional robot logs do not identify which decisions depended on model-imagined futures or how those predictions were resolved. Deployment-time reliability signals include ensemble and verifier scores: ensemble disagreement can regulate imagined rollouts during training~\cite{pan2020m2ac}, and learned verifiers can determine at deployment time whether an imagined trajectory should remain trusted~\cite{wang2026ffdc}. These signals assess the current prediction. A persistent execution record adds a contextual question: under comparable operating conditions and at a comparable prediction horizon, how often has the model's imagination subsequently agreed with reality?

\sys{} addresses this question through a persistent execution record. Each prediction consumed by the planner is registered as a claim and settled when its outcome becomes observable. The result updates a history indexed by operating condition, region, and prediction horizon. An attribution stage excludes outcomes for which model error cannot be separated from localization or sensing error. A lightweight introspection head, trained on settled outcomes with gradients stopped at the world-model backbone, complements sparse or newly encountered histories. The planner combines these sources into credit: sufficient credit permits reliance, while insufficient credit shortens the dependent rollout horizon or triggers observation. Replayable logs retain the prediction, decision-time credit, and eventual outcome.

Here, \emph{where to refuse} refers to the declared operating conditions, regions, and horizons in which current credit falls below the chosen threshold. The partition defines the resolution of the history; execution supplies the evidence within it. \emph{When to refuse} is evaluated before commitment, so new settlements can change a previously permissive decision while the threshold stays fixed. We test placement at matched refusal volume and trace an execution-driven gate transition on hardware. Task completion and verification cost measure the consequences of these interventions.

\textbf{Contributions.} We make two contributions:
\begin{enumerate}
    \item \emph{Execution-settled credit with decision provenance.}
    We connect consumption-defined claims, persistent contextual settlement, and a pre-decision credit query in one deployment record. The gate uses that record to permit reliance or invoke a fallback; replay links each decision to its supporting evidence and eventual outcome (Sec.~\ref{sec:mechanism}).

    \item \emph{An empirical account of where and when history changes refusal.}
    Matched-refusal controls isolate persistent history, randomized audits test denied-candidate risk, and post-warmup shifts test the value of new settlements. Franka trials demonstrate online deployment and a credit-driven gate transition. We report task completion and verification cost alongside these mechanism measures (Sec.~\ref{sec:experiments}).
\end{enumerate}

As a supporting extensibility study, adapters register and settle predictions from unmodified DreamerV3~\cite{hafner2023dreamerv3}, TD-MPC2~\cite{hansen2024tdmpc2}, and V-JEPA 2-AC~\cite{assran2025vjepa2}. They test compatibility across observation, latent, and token representations (Sec.~\ref{sec:mounts}).

\section{Related Work}\label{sec:related}
\sys{} connects per-prediction execution outcomes to contextual credit and decision provenance: claims settle through execution, accumulate in a persistent condition $\times$ region $\times$ horizon file, and support both the next reliance decision and its later audit. The following four bodies of literature supply the estimation, gating, and logging components used to build this deployment record.

\textbf{Instantaneous gating and runtime monitoring.}
After a rollout prefix has been executed, learned verifiers determine whether the remaining imagined actions remain trustworthy~\cite{wang2026ffdc}; pre-execution verification instead vets imagined plans through auxiliary models before commitment~\cite{wu2025foresight}. Anomaly detectors compare predictions with arriving observations and trigger stops or human handovers~\cite{liu2024sirius,liu2024runtime,huang2026failure}. Prediction--observation discrepancies also supply the evidence used by our settlement engine. In model-based RL, ensemble disagreement truncates rollouts during training-time data generation~\cite{janner2019mbpo,pan2020m2ac}, and adaptive variants tune rollout length to local uncertainty~\cite{frauenknecht2024trust}. Our formulation connects these signals to a condition-indexed execution record and per-decision audit: instantaneous signals enter the credit estimator alongside execution history, instantaneous gating remains a baseline family, and a settlement-supervised instantaneous verifier using the same signals serves as the strongest such baseline (Sec.~\ref{sec:gate}). For LLM agents, prediction--observation mismatches have been distilled into heuristic memories that filter low-confidence foresight~\cite{zhang2026worldevolver}; these methods produce a heuristic record rather than settled, condition-indexed statistics.

\textbf{Uncertainty-aware imagination.}
Uncertainty-aware world models penalize unreliable imagined transitions during offline policy training~\cite{yu2020mopo}, explicitly propagate uncertainty along multi-step rollouts on real robots~\cite{li2025rwmu}, or softly truncate visual MPC rollouts at test time using ensemble signals~\cite{du2026elvis}. These signals support training objectives or rollout scheduling; ours governs deployment-time reliance and audit. Calibrated world models attach self-reported uncertainty to generated futures~\cite{mei2025c3}. Our credit estimate combines settled execution history with self-report as a calibrator feature. This grounding is motivated by evidence that internal uncertainty correlates weakly with task risk. Recent work replaces uncertainty penalties with outcome-supervised feedback at planning time~\cite{wang2026feedback}, providing instantaneous grounding without a cumulative, externalized record. Our head follows the confidence-head form~\cite{daftry2016,corbiere2019}, but is supervised by deployment-settled, on-policy outcomes rather than training-set errors. Execution-derived, self-supervised outcome labels have precedents in introspective perception, affordance learning, and online-adapting navigation~\cite{rabiee2019ivoa,kahn2021badgr,mattamala2025wild}; our head differs in its target, labels, and distribution.

\textbf{Certification and competence lines.}
Conformal and robust-control methods study prediction bounds and certified planning, including trust horizons for equivariant world models in synthetic environments~\cite{wang2026certified} and certified robust MPC with learned models on hardware~\cite{li2026wrinkles}. The \sys{} execution history provides an empirical, Mondrian-binned deployment record for both agent and human consumers, evaluated through tier calibration and risk--coverage. Per-bin split-conformal upper bounds on failure remain an optional hardening of the estimator~\cite{vovk2003mondrian}. Competence-aware systems are the closest conceptual predecessors, with each providing part of the required functionality. Learn-from-experience frameworks accumulate place- and appearance-conditioned perception records over repeated traversals and use them to gate autonomy~\cite{gurau2018}, but address a perception module without a prediction-horizon dimension. Introspective perception estimates task-level competence over a deployment map~\cite{rabiee2019ivoa}; subsequent work instead learns location-agnostic error models and projects competence onto the map~\cite{rabiee2022cpip}. Factored self-confidence frameworks combine per-component assessments into reportable machine confidence~\cite{israelsen2025good}. Model-precondition estimators actively collect trajectories to learn where dynamics models support reliable planning~\cite{lagrassa2024}. Where-to-trust estimators learn the spatial reliability of a fixed dynamics model for manipulation planning~\cite{mitrano2021where}, and selective model-based planning learns a model-inadequacy signal from observed errors to determine when planning is beneficial~\cite{abbas2020selective}. Adaptive conformal methods use arriving observations to update uncertainty sets online~\cite{dixit2023}. Our focus is the record linking a consumed prediction, its contextual support, the dependent decision, and its subsequent settlement. Combining these elements requires the predicate and attribution rules in Sec.~\ref{sec:mechanism}. Offline admissibility ladders qualify a simulator before use~\cite{oefinger2026admissibility}; our ledger operates \emph{during} use.

\textbf{Accountability infrastructure and forecast verification.}
Robotics has a tradition of ethical black-box logging~\cite{winfield2017}, and decision-provenance methods trace how decisions propagate through socio-technical systems~\cite{singh2018decision}. Such logs, however, do not capture \emph{reliance on imagination}: which decisions depended on which predictions, the credit assigned at the time, and how those predictions later settled. Meteorology has verified forecasters for decades, including analyses of forecast quality~\cite{murphy1993good} and condition-decomposed skill scores by weather regime~\cite{brier1950,murphy1995}. Intelligence analysis has likewise standardized verbal probability tiers for the human-facing communication problem addressed by our dependency records~\cite{kent1964}. These practices motivate condition-binned agreement records; we turn this offline discipline into an online mechanism within a robot decision loop.

\section{Problem Statement and Scope}

\textbf{System object.}
We consider a learned predictive world model embedded in a robotic decision loop. The model may use any architecture to predict future observations, states, or latent representations over short horizons. \sys{} consumes the model's predictions and, when available, read-only latent features for its introspection head. It does not modify the predictive pathway or training objective of the model.

\textbf{Consumption interface.}
A world model predicts future observations, states, or latents used by a planner. A \emph{claim} freezes a prediction together with its settlement predicate. A reliance decision registers which claim it consumes. Once an outcome is observable and attributable, \emph{settlement} records whether prediction and observation agree. \emph{Credit} estimates the probability of this agreement from contextual history and current reliability signals.

A failed consumed prediction is counted as a \emph{burn}; burn rate is failures per consumed prediction. Verification actions can generate their own claims, whose outcomes are recorded separately. Logs link each prediction use (a \emph{spend}) to its decision-time credit and later settlement. The per-context history is also called the \emph{books}; the figures label the per-decision dependency record a \emph{ticket} and the credit attached to a consumed prediction a \emph{stamp}. Task outcome is \emph{reach} in navigation (the fraction of episodes attaining the goal) and \emph{success} in manipulation (task completion).

\textbf{Evaluation domains.}
We use two co-primary robotics domains. In simulated \emph{aerial navigation}, a quadrotor operates in GPS-denied indoor environments. Its limited sensing range and motion cost create a direct trade-off between relying on imagined free space and acquiring additional observations. In \emph{tabletop manipulation}, evaluated in simulation and on hardware, a Franka manipulator with a wrist-mounted camera predicts occluded or contact-dependent object states. Obtaining additional information may require viewpoint changes or probing actions that incur replanning cost and may disturb the scene. A third 2D navigation domain exercises the same settlement--credit--gating pipeline at low per-episode cost and provides the highest-volume settlement histories for the calibration analysis (Table~\ref{tab:sources}). Across all three domains, only the platform-specific interface (the settlement predicate, condition partition, and available observation action) is re-instantiated. We separately evaluate compatibility with established world-model architectures by mounting the ledger on unmodified external models (Sec.~\ref{sec:mounts}).

\textbf{Scope.}
The object of study is deployment-time reliance on a host model's predictions. The predictive architecture and training objective remain unchanged; settlement, contextual credit, and decision provenance form the added interface. Credit records empirical claim reliability in the environment where it is consumed. Evaluation therefore measures claim calibration, intervention placement, task outcomes, and replayability. Model repair, cross-site transfer, and formal safety certification are separate extensions.

\begin{figure}[t]
\centering
\includegraphics[width=\linewidth]{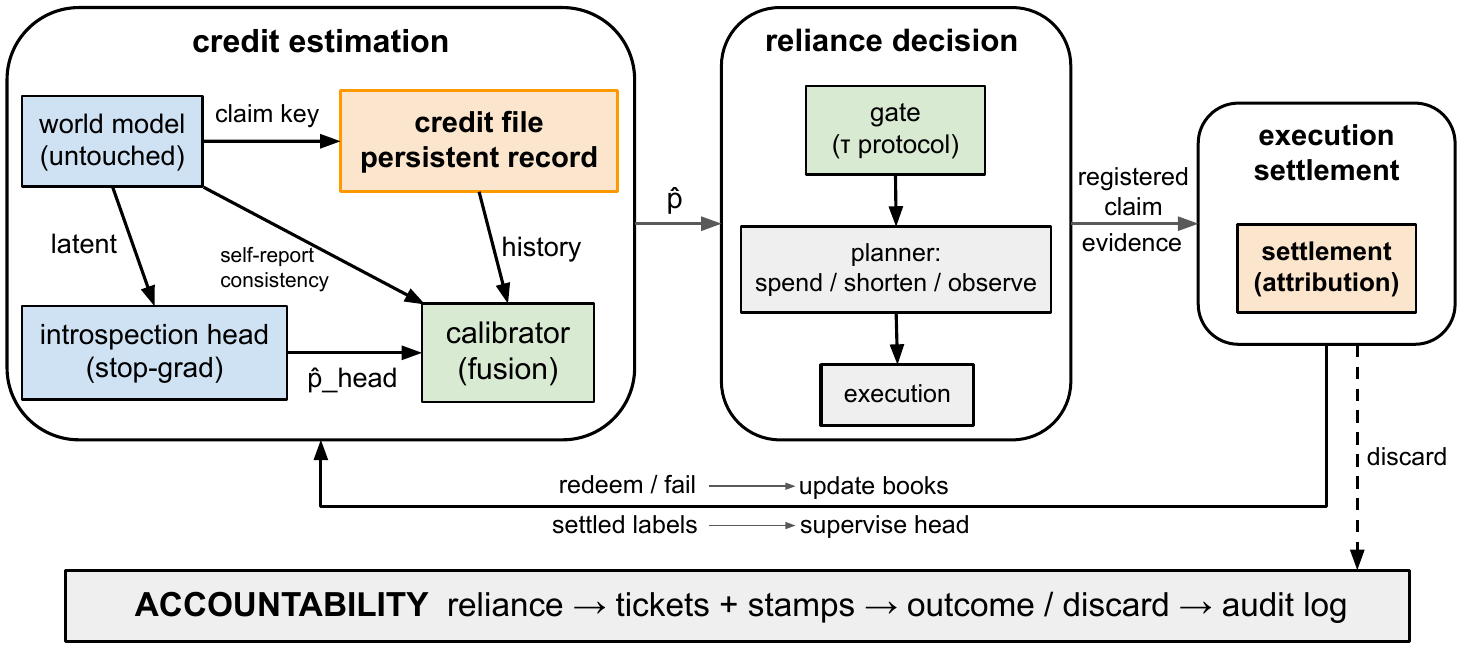}
\caption{\sys{} mechanism. Execution-settled history and model-side signals produce the credit estimate $\hat p$ used to gate reliance. Execution settles registered claims, updates contextual history, and supervises introspection, while both reliance and outcomes remain auditable.}
\label{fig:system}
\end{figure}

\section{Mechanism}\label{sec:mechanism}

\subsection{Settlement engine (execution as settlement)}\label{sec:settlement}

During execution, the model produces short-horizon predictions in predefined horizon buckets. As the corresponding outcomes become observable, consumed predictions are settled against arriving reality without manual labels. A platform-specific \emph{settlement predicate} specifies the compared quantity, agreement threshold, and availability rule. It is declared and frozen before each reported run. For the quadrotor, a predicted occupancy patch beyond the sensing horizon becomes eligible for settlement when the region enters the depth camera's field of view and the state-estimate covariance falls below a threshold. Settlement records agreement or failure; a claim remains pending if its outcome is never observed. For manipulation, a predicted post-push object position is settled when the wrist camera re-observes the object; the claim is an endpoint region sized above the measured aleatoric contact noise.

Across the three domains, we identify five general rules for constructing valid settlement predicates. (1) \emph{Settle the claim consumed by the planner rather than an arbitrary model output}: a probabilistic $|p-\mathrm{obs}|$ score can reward a model that hedges indefinitely near $p=0.5$. (2) \emph{Use typed predicates for heterogeneous claim quantities}: quantities with incompatible units should not share a settlement rule. (3) \emph{Register chained claims only at observable milestones}: intermediate predictions without observable outcomes can distort horizon-dependent histories. (4) \emph{Set the claim tolerance above the aleatoric noise floor}: with approximately $3.4\,\mathrm{cm}$ contact noise, a $2.5\,\mathrm{cm}$ tolerance causes even an otherwise accurate model to fail roughly half of its claims. (5) \emph{Match the settled quantity to the quantity used by the planner}: for endpoint-based manipulation, we settle endpoint regions rather than cumulative motion magnitude. These checks tie agreement to the quantity used for action selection and separate tolerance effects from model reliability.

Attribution separates model disagreement from localization drift and observation noise before an outcome updates credit. We use three stages: (i) \emph{confidence gating}, which permits settlement only under sufficiently reliable state estimates; (ii) \emph{residual decomposition}, which attributes only the component unexplained by the measurement-error model; and (iii) \emph{attribution-based discard}, which excludes unattributable samples and reports their rate. Settlement is necessarily on-policy because only predictions along executed trajectories eventually encounter observable outcomes. The ledger therefore estimates reliability for unexecuted candidates within the same cell under a stated cell-level exchangeability assumption, which we audit in simulation (Sec.~\ref{sec:dose}). Observation actions direct execution toward low-credit regions and generate additional settlements, coupling verification with evidence acquisition.

\subsection{Contextual credit estimation}

Credit estimation starts from the observed agreement of comparable past claims, then combines this history with current reliability signals. Table~\ref{tab:notation} lists the quantities used below.

\begin{table}[t]
\centering
\caption{Notation for settlement and credit. All counts precede the current decision.}
\label{tab:notation}
\footnotesize
\setlength{\tabcolsep}{3pt}
\begin{tabular}{@{}lp{0.69\linewidth}@{}}
\toprule
Symbol & Meaning \\
\midrule
$c_i=(d_i,r_i,k_i)$ & Claim condition, region, and horizon \\
$a_i,\ s_i$ & Attributable outcome available; agreement indicator \\
$N_v,\ S_v$ & Settlements and agreements in context $v$ \\
$q_i,\ n_i,\ W_i$ & Agreement rate, support, and Wilson interval width \\
$h_i,\ u_i,\ \kappa_i$ & Head, self-report, and consistency signals \\
$\hat p_i,\ \tau$ & Estimated agreement probability and gate threshold \\
\bottomrule
\end{tabular}
\end{table}

\emph{Accumulate comparable outcomes.}
A registered claim freezes its prediction, cell $c_i$, predicate $\pi_{c_i}$, and registration time. An available, attributable outcome ($a_i=1$) supplies $s_i=\pi_{c_i}(\hat y_i,y_i)\in\{0,1\}$. For each context, the record stores settlement and agreement counts, together with recency:
\[
\begin{aligned}
N_v&=\sum_{j<i} a_j\mathbf 1[c_j\preceq v],\\
S_v&=\sum_{j<i} a_j s_j\mathbf 1[c_j\preceq v].
\end{aligned}
\]
Here $c_j\preceq v$ means that cell $c_j$ belongs to context $v$. A Mondrian ladder~\cite{vovk2003mondrian} starts at the exact cell and coarsens condition, region, or horizon when support is sparse. The first ancestor with $N_v\ge25$ supplies $q_i=S_v/N_v$, $n_i=N_v$, and Wilson 95\% interval width $W_i=U_v-L_v$. If none is supported, $(q_i,n_i,W_i)=(.5,0,1)$ supplies the history features, allowing the candidate to undergo the ordinary credit test.

\emph{Estimate credit before commitment.}
A logistic calibrator combines contextual history with current signals:
\[
\begin{aligned}
\psi_i^{\rm hist}&=[q_i,\log(1+n_i),W_i],\\
\hat p_i&=\sigma\!\left(w_t^\top[\psi_i^{\rm hist},h_i,u_i,\kappa_i]\right).
\end{aligned}
\]
The interval describes the empirical estimate, not uncertainty in the fused score. Fitting and evaluation are prequential~\cite{dawid1984}: each estimate uses only evidence available before commitment, and later settlements update future estimates. The hardware follow-up uses the empirical rate $\hat p_i=q_i$ directly, as specified before its trials (Sec.~\ref{sec:real}).

\emph{Apply the gate.}
Existing safety logic first avoids predicted-occupied cells and blocks cells without model coverage. For the remaining candidates, $\hat p_i\ge\tau$ permits consumption; otherwise the planner invokes its fallback. The recorded credit stays fixed for audit; dependency records also express it in a fixed five-tier verbal lexicon~\cite{ipcc2010,kent1964}. Observation actions can provide new settlements in a low-credit cell, allowing later decisions to use evidence acquired after earlier refusals.

\subsection{Reliance decisions and audit}

\emph{Agent level:} Before consuming imagination, the planner queries its credit. For a refused candidate, a platform-specific fallback ladder substitutes either a shorter dependent horizon or an observation action. In navigation, the ladder is vantage $\rightarrow$ creep toward the uncertain region through sensor-confirmed space $\rightarrow$ probe by contact. Under a local two-action comparison that prices an observation against a failed consumed prediction, the threshold has the interpretation $p > 1-C_{\text{look}}/C_{\text{burn}}$. This local cost comparison gives a claim-level operating point. To compare reliance policies at a common risk target, the primary experiments select $\tau$ from warmup risk--coverage data and freeze it before evaluation (Sec.~\ref{sec:dose}). Diagnostic snapshots use a separately declared fixed threshold to examine intervention at a common operating point. Section~\ref{sec:supporting} then varies task-failure and collision costs to characterize how these policies translate into task utility. The gate uses the calibrated point estimate $\hat p$. Its conservatism is action-level: low credit may add an observation or shorten reliance on imagination, but cannot remove existing safety behavior. Predicted-occupied cells remain avoided, and unknown cells outside the world model's prediction coverage are blocked. Every committed path cell in navigation therefore belongs to one of five categories: confirmed-free, allowed claim, denied$\rightarrow$shorten-or-look, predicted-occupied$\rightarrow$avoided, or model-uncovered unknown$\rightarrow$blocked.

\emph{Audit level:} The record links a plan to its prediction dependencies, decision-time credit, and supporting settlements. Each subsequent outcome completes the corresponding entry, allowing replay to reconstruct the evidence available at commitment and how the prediction later settled~\cite{liu2026security}.

\subsection{Learning from settled outcomes}

A lightweight head on the model's latent state predicts the agreement probability for each horizon bucket. Settled, on-policy deployment outcomes supervise the head, while gradients are stopped at the backbone to leave the predictive pathway unchanged. The head complements sparse empirical histories: latent features provide additional discrimination in thin or newly encountered cells, whereas dense bins anchor the estimate when direct settlement evidence is abundant. Both sources feed the calibrator, and Sec.~\ref{sec:dose} isolates their contributions through ablation. Holding the predictive pathway fixed isolates changes in credit estimation and reliance from changes in the underlying world model.

\section{Experiments}\label{sec:experiments}

\textbf{Setup.}
We evaluate \sys{} in three robotic domains. \emph{Quadrotor:} The vehicle operates in Isaac Sim indoor corridors ($24\times6$\,m; open, cluttered, and low-texture zones), with depth observations fused into a conservative log-odds belief map. A recurrent latent world model (GRU dynamics over CNN-encoded ego-patches, 1.1M parameters) is trained at three data doses and evaluated over horizon buckets from 0.5 to 5\,s. \emph{Manipulation:} An Isaac Sim Franka uses the official wrist-camera mount. A vision world model (depth+RGB+command $\rightarrow$ displacement, no class labels) is trained on 1{,}280 contact pushes across four material classes. Claims are endpoint regions, and observation actions are viewpoint changes or probing touches. \emph{2D navigation:} A CNN map-completion model operates in procedural lattice worlds. Each episode uses a fresh world (for the primary dose studies: quad 180 across warmup and evaluation, nav2d 90; manipulation: fresh layouts per episode; the persistence tests in Sec.~\ref{sec:gate} report separate counts). The in-domain dose studies hold each architecture fixed to isolate the relationship between training data, prediction quality, and earned credit. External mounts test the settlement interface across architectures (Sec.~\ref{sec:mounts}). The same core stack (settlement, execution history, calibrator, head, gate, and dependency records) runs unchanged across domains and mounts. Only the platform-specific interface (predicate, conditions, and observation action) is re-instantiated. Across the paper, the ledger settles seven claim types, ranging from occupancy cells to token embeddings. Unless stated otherwise, results use three seeds with independent evaluation worlds, and proportions include pooled Wilson 95\% intervals. We use the same prequential estimation and paired evaluation protocol throughout unless explicitly stated otherwise: risk--coverage analysis under an audited exchangeability assumption, credit-source ablations with identical features and labels, and the pre-specified dose design. Threshold protocols are reported for each experiment.

The evaluation addresses four questions: how credit tracks prediction reliability (A), how persistent history changes refusal placement (B), how typed settlement connects different model representations (C), and how the deployed ledger turns new execution outcomes into refusal (D). Supporting analyses quantify the effects of settlement density, observation feedback, partitioning, and task costs.

The evaluation follows the decision loop: estimate credit, locate refusals, and update them as outcomes arrive. Hardware tests deployment and the gate transition; external-model adapters test settlement compatibility. Task costs are assessed separately in Sec.~\ref{sec:supporting}.

\subsection{Credit discrimination and operating points}\label{sec:dose}

\begin{figure}[t]
\centering
\includegraphics[width=0.9\linewidth]{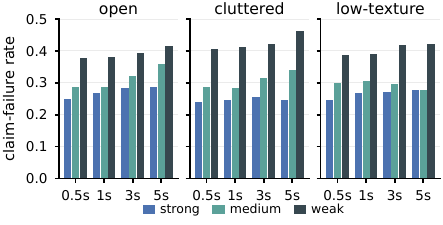}
\caption{Oracle claim failure on the shared held-out distribution. Within each horizon group, bars run from strong $\rightarrow$ weak. Failure is dose-monotone in all 12 cells and generally rises with horizon; the medium/low-texture cell is the exception.}
\label{fig:oracle}
\end{figure}

On a shared held-out state distribution without on-policy selection, claim-failure rates are monotonic in data dose across \emph{all 12} condition$\times$horizon cells (Fig.~\ref{fig:oracle}; quad means: strong $\approx0.26$, medium $\approx0.31$, weak $\approx0.41$). Failure generally increases with horizon: the trend holds in the open and cluttered conditions at every dose, with one reversal for the medium model in the low-texture condition.

The analysis separates model quality from deployment selection. The held-out oracle distribution compares predictions on common states; on-policy histories record the predictions encountered during execution. Comparing the two within matched cells quantifies the exchangeability assumption used to estimate reliability for unexecuted candidates. Across supported condition--horizon cells, the mean within-cell failure-rate gap between executed claims and candidates is $+0.01$/$+0.05$/$+0.10$ for strong/medium/weak models (weak: 90th percentile $+0.18$, worst cell $+0.21$). The candidate-risk gap increases as the model weakens. It limits the interpretation of credit as candidate risk, even when the on-policy score is calibrated.

Two offline analyses of the warmup streams bound the cost of this bias. First, consider a uniform-bias bound in which every credit estimate is optimistic by $\delta$ for unexecuted candidates. Candidate-level failure among consumed predictions retained at $\tau$ is then the on-policy failure of retained predictions plus $\delta$. At the frozen $\tau=0.75$, the on-policy failure rates of retained predictions are $0.177$ (weak), $0.210$ (medium), and $0.217$ (strong). For the weak stream used to select $\tau=0.75$, the pre-stated fail$\leq20\%$ target, defined and met on-policy, therefore transfers to unexecuted candidates only when $\delta\lesssim0.02$. No $\tau$ in $[0.30,0.95]$ restores the target for $\delta\geq0.05$, because the risk--coverage curves bottom out near $0.15$--$0.17$. At the same frozen $\tau$, the medium and strong streams already exceed the target on-policy. Raising the threshold alone therefore does not repair the measured transfer gap.

Second, we apply a measured-gap correction by adjusting the credit of each weak-stream prediction use as $\hat p-\mathrm{gap}(\text{cell})$ using the audited per-cell optimism and re-applying $\tau=0.75$. Coverage falls from $0.27$ to $0.08$ (103 additional denials), yet the newly denied predictions fail on-policy at $0.175$, the same rate as the retained predictions. The correction reduces coverage without enriching the additional denials for observed failure. These measurements quantify the exchangeability assumption required to interpret weak-dose credit as candidate-level risk. The randomized-auditing study in Sec.~\ref{sec:auditshift} complements this oracle comparison by executing sampled denied candidates. It directly tests their failure rate and uses inverse-probability weighting to estimate risk across the settlement-supported candidate stream.

\textbf{Bins, head, and fusion.}
Using identical features and labels, prequential ECE over the 10 comparable (domain, arm) cells in Table~\ref{tab:sources} is lowest for fusion in 6 cells, bins in 3, and the head in 1. When histories are sparse ($n<1000$ settlements), fusion reduces ECE by $0.032$ on average. Dense histories favor the empirical bins at the point estimate (mean bins-minus-fusion ECE gap $=-0.003$; $r=-0.61$ with $\log n$, slope $-0.028$ per decade; a descriptive trend over $n{=}10$ cells, $p\approx0.06$). This ablation identifies the settlement-density regime in which fusion improves estimation; the closed-loop experiments measure its effect on reliance. The persistence battery in Sec.~\ref{sec:gate} separates estimation from memory using a per-cell Beta(1,1) counting baseline without a calibrator, head, or merge ladder. Dense-history results identify persistence as the active ingredient. The post-hoc density ladder in Sec.~\ref{sec:gate} quantifies how, in sparse histories, the full estimator trades additional burn exposure for coverage and reach.

\begin{table}[t]
\centering
\caption{Credit-source ablation using prequential ECE with identical features and labels. Bold indicates the best result in each row; fusion gains concentrate in sparse histories. $^\dagger$ The inference-only mount has no comparable head and is excluded from the 10-cell statistics.}
\label{tab:sources}
\footnotesize
\begin{tabular}{llrrrr}
\toprule
Domain & Arm & $n$ & Bins & Head & Fusion \\
\midrule
nav2d & strong & 10708 & .0252 & \textbf{.0245} & .0284 \\
nav2d & medium & 11580 & \textbf{.0332} & .0363 & .0429 \\
nav2d & weak & 11122 & .0380 & .0365 & \textbf{.0306} \\
quad & strong & 541 & .0746 & .1065 & \textbf{.0620} \\
quad & medium & 561 & .0756 & .0971 & \textbf{.0709} \\
quad & weak & 540 & .0726 & .0988 & \textbf{.0393} \\
manip & strong & 323 & .1349 & .1881 & \textbf{.0636} \\
manip & medium & 315 & .1255 & .1712 & \textbf{.0461} \\
manip & weak & 339 & \textbf{.0728} & .1978 & .0795 \\
TD-MPC2 & walker & 7110 & \textbf{.0192} & .0299 & .0239 \\
V-JEPA 2$^\dagger$ & tabletop & 1223 & \textbf{.0303} & .1129 & .0687 \\
\bottomrule
\end{tabular}
\end{table}

\textbf{Gating at different model doses.}
The quadrotor comparisons characterize gating and pooled credit across model doses. Dense-history tests in Sec.~\ref{sec:gate} isolate persistence. Burn counts measure realized exposure to failed consumed predictions. For the weak quadrotor model (Table~\ref{tab:headline}), credit gating reduces this exposure by 62\% (95\% CI 43--81\%) while retaining the same reach point estimate. Reach is 0.683 in both arms, and the paired difference is $0.00$ with 95\% CI $[-0.17,+0.18]$. The reach interval extends beyond the $\pm0.05$ equivalence margin. We detect no difference in collisions, and episode lengths are similar (495 vs.\ 484 steps). The no-imagination arm never reaches the goal, confirming that the task requires predictions beyond the sensing horizon. Across model doses, the same frozen gate reduces burned imagination by 62\%, 56\%, and 44\% for weak, medium, and strong models, respectively; the respective 95\% CIs are $[43,81]$, $[31,75]$, and $[9,67]$, with the same direction in every seed. The weak collision difference is $-0.8$ ($[-19.2,17.6]$). These contrasts report prediction exposure and task completion at the same operating point. Intervention declines from 10.1 denials per episode for the weak model to approximately 6 for the medium and strong models (full dose$\times$mode results in Table~\ref{tab:dosemode}). 

The response is also dose-graded. In 2D navigation, which uses a fixed, pre-declared $\tau=0.6$ rather than target selection, denials per episode increase from 0 to 13.6 to 14.7 as model dose decreases; observation actions increase from 0 to 1.8 to 1.8, while collisions change from 2.1 to 2.2 to 1.6. In manipulation, consumed predictions per episode decrease from 3.01 to 1.39 to 0.76 (strong/medium/weak) at equal success. In the Isaac vision variant, the strong model can chain at $\tau=0.30$ (3.52 consumed predictions/ep, zero probes), whereas the medium and weak models remain restricted to probe-and-creep behavior.

\begin{table}[t]
\centering
\caption{Weak-dose quadrotor gating versus blind consumption at frozen $\tau=0.75$: 3 seeds $\times$ 20 paired episodes. Burns are totals per 20-episode run, averaged over seeds; other entries are episode means. Brackets are Wilson 95\% CIs for reach.}
\label{tab:headline}
\footnotesize
\resizebox{\linewidth}{!}{%
\begin{tabular}{lcccc}
\toprule
Mode & Reach & Steps & Collisions & Burns \\
\midrule
\textbf{Gated (full)} & \textbf{0.683} [.56,.79] & 495 & 9.85 & \textbf{19.3} \\
Blind consumption & 0.683 [.56,.79] & 484 & 9.08 & 50.3 \\
No imagination & 0.000 [0,.06] & 899 & 0.20 & 0 \\
\bottomrule
\end{tabular}}
\end{table}

\begin{table}[t]
\centering
\caption{Quadrotor dose response at the same frozen $\tau=0.75$: 3 seeds $\times$ 20 paired episodes per dose. Burns are totals per 20-episode run, averaged over seeds; other entries are episode means.}
\label{tab:dosemode}
\footnotesize
\setlength{\tabcolsep}{4.5pt}
\begin{tabular}{llcccc}
\toprule
Dose & Mode & Reach & Steps & Collisions & Burns \\
\midrule
\multirow{2}{*}{Weak} & Gated & 0.683 & 495 & 9.85 & \textbf{19.3} \\
 & Blind & 0.683 & 484 & 9.08 & 50.3 \\
\multirow{2}{*}{Medium} & Gated & 0.750 & 451 & 9.28 & \textbf{25.7} \\
 & Blind & 0.750 & 450 & 8.62 & 58.0 \\
\multirow{2}{*}{Strong} & Gated & 0.683 & 492 & 15.5 & \textbf{26.7} \\
 & Blind & 0.633 & 517 & 9.53 & 47.3 \\
\bottomrule
\end{tabular}
\end{table}

\begin{figure}[t]
\centering
\includegraphics[width=0.68\linewidth]{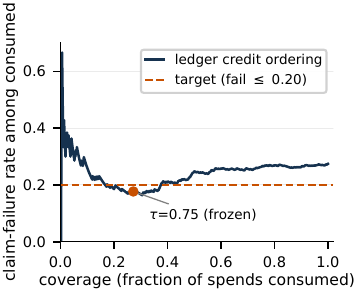}
\caption{Risk--coverage of the calibrated credit $\hat p$ used by the gate, evaluated over 540 consumed predictions settled during warmup for the weak model. The threshold $\tau$ is selected on a 0.05 grid at the pre-stated target and frozen before evaluation. The curve meets the target over the coverage range corresponding to $\tau\in[0.74,0.77]$ (Sec.~\ref{sec:supporting}).}
\label{fig:rc}
\end{figure}

\textbf{Threshold selection.}
The risk--coverage curve from 540 consumed predictions settled during warmup (overall claim failure 27.4\%) yields $\tau=0.75$ at the pre-stated fail$\leq20\%$ target (Fig.~\ref{fig:rc}); this threshold is then frozen for evaluation. An earlier, manually selected $\tau=0.6$ exhibited three failure modes (pinch-point blindness, invisible chokepoints, and no drift response) that disappear at the protocol-selected operating point. Section~\ref{sec:supporting} examines sensitivity to the thresholded score and sample size. The closed-loop outcome across seeds provides the final validation (Table~\ref{tab:headline}).

Because $\tau$ also determines the margin at which drift affects consumption (Sec.~\ref{sec:supporting}), threshold selection and drift sensitivity are coupled design choices.

\textbf{Against instantaneous gates.}
We compare against ensemble disagreement (M2AC-style), self-report thresholding, one-step consistency monitoring, and sliding-window adaptive conformal prediction using the same model (strong dose), worlds, and warmup data. We also include a settlement-supervised \emph{book-blind verifier}: it retains the same settlement stream, calibrator, and introspection head, but its credit estimate omits the three persistent-history features (Table~\ref{tab:named}). This isolates persistent history from settlement supervision without disabling settlement or audit.

The quadrotor protocol with sparse histories separates score ranking from operating-point selection. In the original three-seed sample, the book-blind verifier had lower area under the risk--coverage curve (AURC; lower is better) than the ledger ($.249\pm.067$ vs.\ $.356\pm.055$). A post-hoc replication with settlement streams did not preserve this separation ($.269\pm.031$ vs.\ $.272\pm.082$, with mixed paired directions); over the six paired runs, the ledger--book-blind AURC difference is $+.055$ with a per-run $t$-interval $[-.034,+.144]$. The pooled interval leaves the global-ranking difference unresolved, while operating-point stability replicates. Across the two samples, the ledger and book-blind verifier select moderate target-risk thresholds ($.53$--$.71$ and $.74$--$.78$), whereas raw instantaneous signals again produce boundary or unstable values ($\tau^*\approx1$, $\approx.22$, or large seed-to-seed variation). The original closed-loop sample measures the associated operating trade-offs: burns span 9.0--42.7 and reach spans 0.55--0.80 across arms and seeds; adaptive conformal prediction issues only 3.4 denials/episode but produces the most burns.

Persistent history matters when the location of verification is consequential. In manipulation (Table~\ref{tab:manipnamed}), disabling history features increases probes from 0.36 to 1.00/episode in every seed, at success 0.94 [0.88,0.97] versus 0.98 [0.94,1.00]. These measurements characterize the trade-off between verification and reliance. The pre-registered 10-seed battery reproduces the direction (Table~\ref{tab:manippersist}): probes 0.55 vs.\ 1.00 at success 0.90 vs.\ 0.93. The two studies use different seeds and compute platforms; the battery is the pre-registered result. The hardware experiments separately trace online credit updates and auditable provenance (Sec.~\ref{sec:real}), linking changed credit to the resulting decision.

\begin{table}[t]
\centering
\caption{Matched quadrotor gates for the strong model. Top: AURC (mean$\pm$sd over three seeds) in the original sample and a fresh post-hoc replication, comparing the reproducibility of score ranking. Bottom: original closed-loop operating points; each $\tau^*$ meets fail$\leq20\%$ on warmup, and the last two columns average each seed's total over 20 episodes.}
\label{tab:named}
\footnotesize
\setlength{\tabcolsep}{4pt}
\begin{tabular}{lcc}
\toprule
Arm & Original AURC$\downarrow$ & Replication AURC$\downarrow$ \\
\midrule
\textbf{Ledger (ours)} & .356\,$\pm$\,.055 & .272\,$\pm$\,.082 \\
Book-blind verifier & .249\,$\pm$\,.067 & .269\,$\pm$\,.031 \\
Ensemble (5 members) & .285\,$\pm$\,.097 & .282\,$\pm$\,.107 \\
Consistency monitor & .284\,$\pm$\,.039 & .311\,$\pm$\,.052 \\
Self-report & .309\,$\pm$\,.060 & .251\,$\pm$\,.057 \\
Adaptive conformal & .263\,$\pm$\,.079 & .296\,$\pm$\,.099 \\
\bottomrule
\end{tabular}
\vspace{2pt}

\begin{tabular}{lccc}
\toprule
Arm & Original $\tau^*$ (s0/s1/s2) & Denials/ep & Burns \\
\midrule
\textbf{Ledger (ours)} & \textbf{.58\,/\,.66\,/\,.64} & 7.8 & 24.3 \\
Book-blind verifier & .78\,/\,.75\,/\,.74 & 14.2 & 15.3 \\
Ensemble (5 members) & 1.0\,/\,.68\,/\,1.0 & 12.6 & 24.0 \\
Consistency monitor & .23\,/\,.23\,/\,.22 & 15.4 & 13.7 \\
Self-report & 1.0\,/\,1.0\,/\,1.0 & 19.3 & 9.0 \\
Adaptive conformal & 1.0\,/\,.30\,/\,1.0 & 3.4 & 42.7 \\
\bottomrule
\end{tabular}
\end{table}

\subsection{Persistent history and refusal placement}\label{sec:gate}

\textbf{Isolating persistent history.}
The book-blind verifier omits persistent-history features from credit estimation while retaining the settlement-supervised, prequential calibrator, head, and settlement record. The domains provide complementary support regimes. In the quadrotor protocol, most condition$\times$region$\times$horizon cells are pooled below the merge threshold (Sec.~\ref{sec:limits}). Dense 2D-navigation histories provide the resolution needed to compare persistent cell histories, using a snapshot at a declared operating point and a pre-registered battery over the full operating frontier.

The snapshot uses the medium model and the domain's fixed, pre-declared $\tau=0.6$. The fixed threshold defines the snapshot operating point; the frontier and target-selected comparisons below evaluate sensitivity to that choice. The experiment comprises 30 blind warmup episodes followed by 30 gated episodes on world seeds shared across arms; the warm histories are carried into a fresh calibrator and head, with 3 seeds in total (Table~\ref{tab:persist}). Two ablations remove history at different stages. \emph{Warm-only} histories are restored to the warmup snapshot at the beginning of every episode, whereas \emph{episode-local} histories are cleared at the beginning of every episode. Removing persistence does not change the model's errors: the failure rate over the first 25 consumed predictions of each episode is $0.26$ in every arm. It does, however, change where the gate intervenes. With persistent histories, 69\% of denials occur in cells with prior failures. Episode-local books instead place three times as many denials in the two healthy conditions (349 vs.\ 117 per 30 episodes; higher in 3/3 seeds) and incur 28\% more burns (1287 vs.\ 1004 per 30 episodes; 3/3 seeds). Equivalently, persistence reduces burns by 22\%, with a hierarchical-bootstrap 95\% CI of $[8,33]$. The burn reduction accompanies reach of 0.70 for persistent histories versus 0.76 for episode-local books. Warm-only books and the full ledger have comparable outcomes in this snapshot, locating the persistence effect in the retained warmup evidence.

The effect becomes stronger under localized recurrent failure. Adding 15\% more obstacles to the cluttered third of every world, during both warmup and evaluation, increases the settled failure rate for that condition from $0.33$ to $0.42$. The model, trained at nominal density, continues to predict free space in that region. The ledger concentrates denials on the degraded condition (82\% in previously failed cells; 90 healthy-condition denials per 30 episodes vs.\ 1232 for the book-blind verifier), consumes 46\% fewer failing claims there (624 vs.\ 1168; 3/3 seeds), and incurs 53\% fewer burns (693 vs.\ 1489; 95\% CI $[41,66]$). Healthy-condition denials decrease by 93\% (90 vs.\ 1232; bootstrap CI $[55,96]$). This reduction comes at the cost of task completion (0.29 vs.\ 0.39), because the ledger substitutes observation for imagination in the degraded region (7.9 vs.\ 3.2 observation actions per episode). This snapshot links the ledger's response to previously observed failures and quantifies its observation--completion trade-off. The matched-refusal battery below isolates placement at equal refusal volume.

\begin{figure}[t]
\centering
\includegraphics[width=\linewidth]{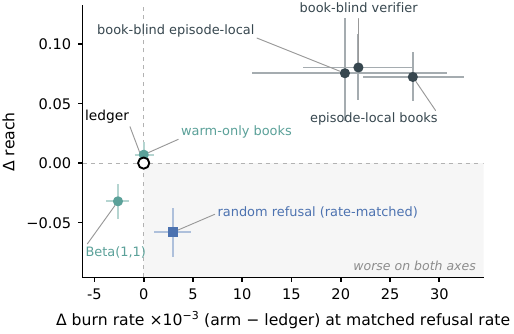}
\caption{Refusal placement at matched refusal rate (2D navigation, 10 seeds; Table~\ref{tab:matched}). Points are paired arm-minus-ledger means. Horizontal and vertical error bars show pooled bootstrap 95\% CIs for burn rate and reach, respectively; thin slanted lines connect labels. The shaded quadrant has higher burn rate and lower reach. Seed-level checks are reported in the text.}
\label{fig:matched}
\end{figure}

\textbf{Matched-conservativeness battery.}
The persistence comparison isolates three effects. Matching refusal volume controls how much imagination is withheld; a rate-matched \emph{random} gate tests whether history selects different claims; and a minimal counting baseline separates persistent evidence from the additional estimation components. A denied prediction use triggers the same observation ladder in every arm, so matching refusal rates also matches the intervention mechanism. A pre-registered 610-run main battery includes all three, with grids, endpoints, and pass criteria frozen before data collection (medium model, 10 seeds, $\tau\in[0.30,0.90]$ in steps of $0.05$); a separate 90-run warmup-density ladder is used below. The battery completes a $2\times2$ factorial ablation \{history features on, off\}$\times$\{persistent, episode-local\}. It also includes an i.i.d.\ random-denial gate swept over $p$, such that its realized refusal rate brackets that of the ledger, and a per-cell Beta(1,1) counting baseline without a calibrator, head, or merge ladder. Because suppressing consumed predictions mechanically reduces absolute burns, every comparison reports two accounting conventions: absolute burns per episode and the volume-normalized \emph{burn rate}, defined as burns per consumed prediction (this convention was added post hoc; see Appendix~A). We compare arms at matched conservativeness. For each seed, an arm's realized refusal rate is located on the ledger's own $\tau$-parameterized frontier, and paired differences are pooled over the grid (Table~\ref{tab:matched}; Fig.~\ref{fig:matched}). Because points on the grid share seeds, we also aggregate matched differences within seed and compute $t$-intervals. These checks preserve every burn-rate and reach conclusion; the random burn-rate interval and the book-blind episode-local absolute-burn interval include zero.

The battery yields three results. First, every ablation of history features or persistence has a higher burn rate at the same refusal volume. Their pooled absolute-burn differences are also positive, although the seed-level interval for the combined book-blind, episode-local arm crosses zero. Episode-local books consume slightly \emph{less} yet still burn more, so the main effect is not an artifact of consumption volume. Moreover, their absolute-burn penalty increases with settlement density (warmup 3/10/30/60 episodes: $+7.6$, $+9.9$, $+10.2$, $+17.2$ burns/ep; pooled CIs exclude zero). These arms recover task success ($+0.07$ to $+0.08$ reach), reproducing the reach-for-burns trade-off in Table~\ref{tab:persist}.

Second, random refusal changes consumption volume and task reach even after matching refusal rates. It consumes 56 fewer predictions per episode (CI $[-70,-43]$), making its \emph{absolute} burns appear competitive ($-0.6$, CI includes zero), but yields lower reach ($-0.06$; pooled bootstrap CI $[-0.08,-0.04]$, per-seed $t$-interval $[-0.09,-0.02]$, lower in 9/10 seeds). Its burn rate is directionally higher ($+0.003$; pooled CI $[+0.001,+0.005]$), while the per-seed $t$-interval $[-.0003,+.0062]$ includes zero.

Third, warm-only books are indistinguishable from the full ledger, and Beta(1,1) counting lies on the same trade-off frontier with dense histories (burn rate $-0.003$ at reach $-0.03$). These controls locate the placement effect with dense histories in retained evidence, with counting and fusion occupying different points on the reach--burn-rate frontier.

A post-hoc Beta density ladder qualifies the case of sparse histories. With only 3 or 10 warmup episodes, raw per-cell counting refuses much more often than the full estimator (about 63--67 vs.\ 47--48 denials/episode). It consequently burns 6--9 fewer claims but loses 9--12 percentage points of reach. Thus, the calibrator, head, and merge ladder buy coverage rather than lower burn exposure when histories are sparse; whether that trade is useful depends on the cost of task failure. The ECE ablation in Table~\ref{tab:sources} and the density ladder thus connect estimation quality to the coverage available to the planner.

The protocol-selected threshold produces the same result at the dose that requires gating. Under the fail$\leq20\%$ target, the weak model (dose $0.02$) selects an interior $\tau^*=0.709\pm0.036$, and the gate remains active in 10/10 seeds for every arm. The ledger incurs 8.6 burns per episode, compared with 15.3 for episode-local books and 20.2 for the book-blind verifier. At the medium dose, the protocol selects $\tau^*\approx0.38$ and intervenes in only 2/10 seeds. This reproduces the earlier observation that target selection stands down when the model is sufficiently reliable, motivating the fixed $\tau=0.6$ used for the preceding diagnostics. A more severely corrupted regime tests the operating boundary: task completion for every arm collapses to $\approx0$, and random refusal maintains a burn-rate advantage ($-0.028$). In dense histories, Beta(1,1) also retains its slight burn-rate advantage at a small reach cost. These regimes distinguish the value of persistent evidence from the choice of estimator and the availability of a viable task policy.

\textbf{Persistence in manipulation.}
Repeating the battery in manipulation (10 seeds, protocol-selected $\tau^*$ per arm; Table~\ref{tab:manippersist}, top) provides the clearest task-level dependence on persistence, although its magnitude is modest. Resetting histories at each episode reduces success from 0.90 to 0.73 and restores the probe cost to $\approx1$ per episode, because cross-episode class credit is required to license chained consumption. The low absolute burn count of episode-local books (1.3 vs.\ 16.5) reflects near-total abstention from consumption (0.26 vs.\ 4.46 consumed predictions/ep), not better refusal placement. Against the book-blind verifier, the ledger instead trades success (0.90 vs.\ 0.93) for fewer probes (0.55 vs.\ 1.00/episode). These volume and verification effects motivate reporting burns together with burn rate.

The original random-$p$ grid bracketed the ledger's refusal rate in only 9/10 seeds and estimated a $+5.4$-percentage-point ledger success difference. Its bootstrap interval is $[1.1,9.6]$ points and its seed-level $t$-interval is $[0.02,10.7]$ points; the expanded analysis below addresses the incomplete seed coverage. The omitted seed was the ledger's lowest-success run, so we froze a targeted post-hoc extension before collecting additional data: $p\in\{0.9,0.95\}$ completed coverage for the original seeds, and the ledger protocol point plus the full random-$p$ sweep were repeated on independent seeds 10--19. Including the omitted original seed reduces that block's difference to $+4.8$ points (seed-level 95\% $t$-interval $[-0.2,9.7]$). The independent replication preserves the direction at $+2.9$ points (7/10 seeds) but does not exclude zero ($[-0.8,6.7]$), leaving the independent success difference unresolved under the pre-specified analysis rule. Pooled over all 20 seeds, the ledger difference is $+3.8$ points (bootstrap interval $[1.4,6.5]$, seed-level interval $[1.0,6.7]$; 14/20 seeds). The refusal-placement result is stronger: random refusal has a higher burn rate both in the replication alone ($+.048$, seed-level interval $[.008,.087]$) and pooled ($+.043$, $[.013,.073]$), and triggers $+.276$ probes/episode in the pooled comparison ($[.206,.345]$; Table~\ref{tab:manippersist}, bottom). Together, the independent and pooled analyses support a replicated placement advantage, with a task-success gain resolved in the pooled sample.

\textbf{What the placement results establish.}
Persistent credit changes which claims the gate consumes. In navigation, the book-blind verifier and episode-local books attain higher reach at higher burn rate; in manipulation, the independent random-refusal comparison resolves higher burn rate and probe use, while the success difference is resolved only after pooling both samples. These are different operating trade-offs, assessed through task costs in Sec.~\ref{sec:supporting}. The next study tests denied-candidate risk directly and then asks whether settlements collected after warmup improve adaptation.

\begin{table}[t]
\centering
\caption{Effect of persistence (2D navigation, medium model, frozen $\tau=0.6$, 3 seeds $\times$ 30 gated episodes; per-seed totals averaged across seeds). Healthy den.: denials in the two non-cluttered conditions; Targeted: fraction of denials in cells with a prior failure; Rate: pooled burns per consumed prediction. Episode length varies with reach, so consumed predictions/ep is not implied by denials/ep. Bottom: the same arms with 15\% additional obstacles in the cluttered condition.}
\label{tab:persist}
\footnotesize
\setlength{\tabcolsep}{4pt}
\resizebox{\linewidth}{!}{%
\begin{tabular}{lcccccc}
\toprule
Arm & Reach & Denials/ep & Healthy den. & Targeted & Burns & Rate \\
\midrule
\textbf{Ledger (persistent books)} & 0.70 & 20.1 & \textbf{117} & \textbf{.69} & \textbf{1004} & .075 \\
Warm-only books & 0.68 & 24.4 & 136 & .78 & 961 & .067 \\
Episode-local books & 0.76 & 29.5 & 349 & .18 & 1287 & .094 \\
Book-blind verifier & 0.78 & 8.6 & 137 & .27 & 1333 & .096 \\
\midrule
\multicolumn{7}{l}{\emph{Localized recurrent failure (cluttered $+15\%$ obstacles)}} \\
\textbf{Ledger (persistent books)} & 0.29 & 134 & \textbf{90} & \textbf{.82} & \textbf{693} & \textbf{.052} \\
Book-blind verifier & 0.39 & 96 & 1232 & .19 & 1489 & .083 \\
\bottomrule
\end{tabular}}
\end{table}

\begin{table}[t]
\centering
\caption{Matched-conservativeness battery (2D navigation, medium model, 10 seeds $\times$ $\tau\in[0.30,0.90]$). Entries are paired differences, arm minus ledger, at the same realized refusal rate on the ledger's frontier, pooled over the grid; intervals are bootstrap 95\% CIs. The per-seed $t$-interval includes zero for $^\dagger$ burn rate and $^\ddagger$ absolute burns.}
\label{tab:matched}
\footnotesize
\setlength{\tabcolsep}{3pt}
\resizebox{\linewidth}{!}{%
\begin{tabular}{lccc}
\toprule
Arm vs.\ ledger & $\Delta$Burns/ep & $\Delta$Burn rate & $\Delta$Reach \\
\midrule
Book-blind verifier & $+9.3$ {\scriptsize$[6.9,11.7]$} & $+.022$ {\scriptsize$[.016,.027]$} & $+.08$ {\scriptsize$[.05,.11]$} \\
Episode-local books & $+10.2$ {\scriptsize$[8.3,12.1]$} & $+.027$ {\scriptsize$[.022,.033]$} & $+.07$ {\scriptsize$[.05,.09]$} \\
Book-blind, episode-local$^\ddagger$ & $+3.8$ {\scriptsize$[1.6,6.4]$} & $+.020$ {\scriptsize$[.011,.031]$} & $+.08$ {\scriptsize$[.04,.12]$} \\
Warm-only books & $+0.0$ {\scriptsize$[-0.5,0.6]$} & $+.000$ {\scriptsize$[-.001,.001]$} & $+.007$ {\scriptsize$[-.004,.018]$} \\
Beta(1,1) counts & $-1.5$ {\scriptsize$[-2.0,-0.9]$} & $-.003$ {\scriptsize$[-.004,-.001]$} & $-.03$ {\scriptsize$[-.05,-.02]$} \\
Random (rate-matched)$^\dagger$ & $-0.6$ {\scriptsize$[-1.5,0.3]$} & $+.003$ {\scriptsize$[.001,.005]$} & $-.06$ {\scriptsize$[-.08,-.04]$} \\
\bottomrule
\end{tabular}}
\end{table}

\begin{table}[t]
\centering
\caption{Manipulation persistence and targeted replication. Top: original 10-seed battery, 40 episodes/seed at protocol-selected $\tau^*$; the random row covers 9/10 seeds. Rate is burns per consumed prediction. Bottom: $\Delta$ success is ledger minus random (percentage points); other deltas are random minus ledger. Brackets are seed-level 95\% $t$-intervals.}
\label{tab:manippersist}
\footnotesize
\setlength{\tabcolsep}{3pt}
\resizebox{\linewidth}{!}{%
\begin{tabular}{lcccccc}
\toprule
Arm & Success & Denials/ep & Probes/ep & Uses/ep & Burns & Rate \\
\midrule
\textbf{Ledger (persistent books)} & 0.90 & 3.52 & 0.55 & 4.46 & 16.5 & .092 \\
Random (rate-matched) & 0.85 & 3.52 & 0.85 & 4.11 & 17.4 & .106 \\
Book-blind verifier & 0.93 & 1.38 & 1.00 & 6.52 & 28.9 & .111 \\
Episode-local books & 0.73 & 7.03 & 0.97 & 0.26 & 1.3 & .124 \\
Beta(1,1) counts & 0.86 & 3.06 & 0.52 & 4.62 & 15.6 & .085 \\
\bottomrule
\end{tabular}}
\vspace{2pt}

\resizebox{\linewidth}{!}{%
\begin{tabular}{lcccc}
\toprule
Block & Covered & $\Delta$Success (pp) & $\Delta$Rate & $\Delta$Probes/ep \\
\midrule
Original grid & 9/10 & $+5.4$ $[+0.02,10.7]$ & $+.016$ $[-.007,.040]$ & $+.306$ $[.194,.417]$ \\
Original + coverage & 10/10 & $+4.8$ $[-0.2,9.7]$ & $+.038$ $[-.015,.091]$ & $+.275$ $[.155,.395]$ \\
Independent replication & 10/10 & $+2.9$ $[-0.8,6.7]$ & $+.048$ $[.008,.087]$ & $+.276$ $[.180,.373]$ \\
Pooled & 20/20 & $+3.8$ $[1.0,6.7]$ & $+.043$ $[.013,.073]$ & $+.276$ $[.206,.345]$ \\
\bottomrule
\end{tabular}}
\end{table}

\begin{table}[t]
\centering
\caption{Manipulation baselines (3 seeds $\times$ 40 episodes; Success includes Wilson 95\% CIs). The comparison measures verification effort alongside task success under each reliance policy.}
\label{tab:manipnamed}
\footnotesize
\setlength{\tabcolsep}{4pt}
\resizebox{\linewidth}{!}{%
\begin{tabular}{lcccc}
\toprule
Arm & $\tau^*$ & Success & Probes/ep & Actions \\
\midrule
\textbf{Ledger (ours)} & \textbf{0.553} & 0.94 [.88,.97] & \textbf{0.36} & \textbf{9.6} \\
Book-blind verifier & 0.739 & 0.98 [.94,1.0] & 1.00 & 10.2 \\
Self-report & 0.970 & 1.00 [.97,1.0] & 1.00 & 10.1 \\
Adaptive conformal & 0.970 & 1.00 [.97,1.0] & 1.00 & 10.1 \\
Ensemble (5 members) & 0.987 & 0.79 [.71,.85] & 0.64 & 10.7 \\
\bottomrule
\end{tabular}}
\end{table}

\subsection{Auditing denials and adapting to a shift}\label{sec:auditshift}

\textbf{Randomized auditing of denials.}
To test the risk behind a refusal directly, we randomly execute a fraction of the candidates that the gate would deny. Each audit executes the original candidate, allowing its own prediction to settle. The ledger's update rules and the fallback for unaudited denials remain unchanged. This post-hoc study fixes its design before collection: medium model, $\tau\in\{0.5,0.6,0.7\}$, audit probability $\varepsilon\in\{0.2,0.5\}$, and 10 seeds (Table~\ref{tab:auditshift}).

In the healthy regime, audited candidates fail at $0.31$--$0.34$ versus $0.17$--$0.21$ for allowed candidates at $\tau=0.6$ and $0.7$. The paired difference is $+0.10$ to $+0.17$, with bootstrap CIs excluding zero and the same direction in 9--10/10 seeds at both audit probabilities. Across the threshold grid under localized degradation, the rates separate further: $0.50$--$0.77$ versus $0.19$--$0.21$ ($\Delta=+0.30$ to $+0.56$, positive in 10/10 seeds). The lower refusal volume at $\tau=0.5$ in healthy worlds supplies 12--27 audits per run and correspondingly wider intervals. Randomized execution thus directly tests whether denied candidates fail more often on the audited decision stream, without imputing their outcomes from within-cell exchangeability.

Reweighting audited settlements by $1/\varepsilon$ also estimates failure across allowed and denied candidates. At the two active healthy operating points, this inverse-probability-weighted estimate is $0.21$--$0.24$, exceeding the allowed-only rate by $0.01$--$0.05$; under degradation it is $0.31$--$0.33$, about $0.12$ above the allowed-only rate. This direction and scale agree with the selection gap observed in the oracle audit (Sec.~\ref{sec:dose}). The estimate concerns settlement-supported candidates encountered under auditing (Appendix~A). Weighting applies to this evaluation; online credit updates follow the original rules. Executing the original candidate makes the simulation audit complementary to hardware probes, which settle a substitute action's prediction (Sec.~\ref{sec:real}).

\textbf{Online settlement under a post-warmup shift.}
The stationary comparisons in Table~\ref{tab:matched} show that warmup history can remain useful throughout evaluation. To isolate the contribution of later settlements, a second post-hoc study introduces additional obstacles in the cluttered condition only after 30 nominal warmup episodes, followed by 30 gated episodes. Each new-run design is frozen before collection, using the same medium model, three thresholds, and 10 seeds. Warm-only books return to the warmup snapshot at each episode start, while their head and calibrator continue learning online. The comparison therefore isolates carrying new settlement evidence across episodes from the estimator updates shared by both arms.

At the $+15\%$ obstacle level of Table~\ref{tab:persist}, warm-only books incur $2.3$ more burns per episode than the online ledger ($[0.8,4.0]$; Table~\ref{tab:auditshift}). Their burn-rate difference is smaller ($+0.002$, $[0.000,0.004]$), and the reach difference is $+0.013$ ($[-0.008,0.036]$). Under the severe shift ($q=1$, completion $\approx0$ in every arm), the warm-only minus online burn-rate difference is $+0.014$ ($[0.010,0.018]$), pooled across thresholds; it is positive at all three operating points. The moderate shift thus resolves a reduction in consumed failures, while the severe shift gives the clearest separation in failures per consumed prediction.

The time course illustrates how the record changes intervention. At $+15\%$ obstacles and $\tau=0.6$, the online ledger's burns per episode in the affected condition fall from $27.6$ to $19.7$ between the first and last ten episodes, as denials per episode rise from $57$ to $133$. Warm-only burns fall from $25.3$ to $22.9$, alongside continued head and calibrator updates. Book-blind and episode-local burns rise from $38.8$ to $52.4$ and $33.2$ to $48.4$, respectively; their higher overall reach ($+0.10$ and $+0.07$) characterizes the accompanying trade-off. Together, these comparisons identify the additional adaptation supplied by persistent history updates and its dependence on shift severity and operating point.

\begin{table}[t]
\centering
\caption{Randomized auditing and post-warmup adaptation (2D navigation, medium model, 10 seeds). Top: failure rates of audited and allowed candidates; $\Delta$ is audited minus allowed. Bottom: arm minus online-ledger differences, averaged over $\tau\in\{0.5,0.6,0.7\}$. Brackets are bootstrap 95\% CIs over seeds (top) or paired seed--threshold contrasts (bottom).}
\label{tab:auditshift}
\footnotesize
\setlength{\tabcolsep}{3pt}
\resizebox{\linewidth}{!}{%
\begin{tabular}{llccc}
\toprule
\multicolumn{5}{l}{\emph{Randomized auditing ($\varepsilon=0.2$; $\varepsilon=0.5$ agrees)}} \\
Regime & $\tau$ & Audited fail & Allowed fail & $\Delta$ [95\% CI] \\
\midrule
Healthy & 0.6 & .311 & .208 & $+.103$ [.058,.139] \\
Healthy & 0.7 & .341 & .188 & $+.154$ [.135,.171] \\
Degraded & 0.6 & .719 & .203 & $+.516$ [.489,.541] \\
Degraded & 0.7 & .527 & .194 & $+.334$ [.308,.356] \\
\midrule
\multicolumn{5}{l}{\emph{Post-warmup shift, $+15\%$ obstacles (arm $-$ online ledger)}} \\
Arm & & $\Delta$Burns/ep & $\Delta$Burn rate & $\Delta$Reach \\
\midrule
Warm-only books & & $+2.3$ [0.8,4.0] & $+.002$ [.000,.004] & $+.013$ [$-$.008,.036] \\
Beta(1,1) counts & & $+6.5$ [4.1,9.0] & $+.009$ [.006,.012] & $+.032$ [.004,.060] \\
Episode-local books & & $+18.3$ [13.4,23.3] & $+.036$ [.029,.043] & $+.072$ [.038,.108] \\
Book-blind verifier & & $+21.8$ [15.8,27.9] & $+.041$ [.032,.053] & $+.099$ [.066,.132] \\
\bottomrule
\end{tabular}}
\end{table}

\subsection{Settlement across world-model representations}\label{sec:mounts}

\begin{figure}[t]
\centering
\includegraphics[width=\linewidth]{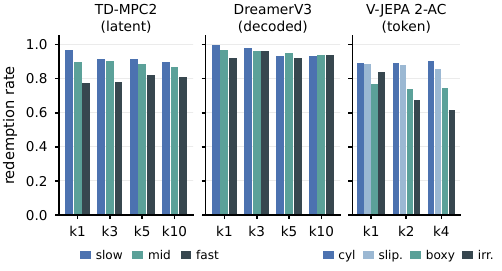}
\caption{Contextual credit for three external world-model hosts. Settlement reveals host-specific reliability, including speed dependence in the RL models and object-class dependence for V-JEPA 2-AC on an unseen embodiment, without providing class supervision to the world model.}
\label{fig:mounts}
\end{figure}

We mount the same ledger on three established open-source world models spanning recurrent, decoder-free latent, and foundation-scale video-representation interfaces. Each mount requires only environment pinning and an approximately 200-line adapter (Table~\ref{tab:hosts}, Fig.~\ref{fig:mounts}).

\emph{TD-MPC2 (latent settlement):}
Using the official walker-walk checkpoint, we define claims as latent predictions over executed actions and freeze $\theta_k$ at the 90th percentile of one calibration episode for each $k\in\{1,3,5,10\}$. Across 7{,}110 settlements in six episodes, agreement rate varies with movement speed at every horizon ($k{=}1$ slow/mid/fast $=$ 0.966/0.898/0.776). The ordering reverses on cheetah-run: claims in the slow regime agree with observations at rates of 0.68--0.77, compared with 0.78--0.83 in the mid/fast regimes. The ledger therefore records host- and task-specific reliability rather than imposing a fixed ordering.

\emph{DreamerV3 (decoded-observation settlement):}
Using the official walker-walk recipe, we decode imagination over executed actions through the official open-loop path. Across 23{,}910 settlements in six episodes, agreement rate in the slow regime decreases from 0.998 at $k=1$ to 0.930 at $k=10$; at $k=1$, rates also vary with speed.

\emph{V-JEPA 2-AC (token-space settlement):}
We use the internet-pretrained, action-conditioned ViT-g model~\cite{assran2025vjepa2} for inference only on our Isaac tabletop, an embodiment outside its post-training distribution. Across 1{,}223 settlements from 16 held-out trajectories, agreement rates vary with commanded action magnitude at every horizon ($k=1$ small/mid/large $=$ 0.91/0.93/0.78; $k=4$: 0.77/0.83/0.63). A matched collection with object-class labels (1{,}175 settlements; Fig.~\ref{fig:mounts}, right) shows high agreement rates for cylinders and sliders ($\sim$0.86--0.90) and the strongest horizon decay for irregular objects (0.84$\rightarrow$0.62). Settling the same checkpoint against real Franka execution frames ($k=1$, 240 settlements) preserves the rank ordering of the four classes (1.00/0.98/0.95/0.60). Raw token distances remain consistently ordered, while the calibrated tolerance widens from $\theta=22.0$ to 32.2. The same settlement abstraction therefore applies to both simulated and real observations without modifying the foundation model.

\begin{table}[t]
\centering
\caption{A common ledger across world-model hosts. The settlement space and claim predicate define the host-specific interface.}
\label{tab:hosts}
\setlength{\tabcolsep}{3.5pt}
\footnotesize
\resizebox{\linewidth}{!}{%
\begin{tabular}{llll}
\toprule
Host (world model) & Settlement space & Predicate & Reliability varies by \\
\midrule
Quad GRU-RSSM & occupancy patch & cell claim & dose, horizon \\
Nav2d CNN completion & occupancy map & cell claim & dose, condition \\
Manip vision WM & object endpoint & region claim & material class \\
TD-MPC2 (official) & latent $z$ & $\|\hat z{-}z^*\|{<}\theta_k$ & speed, horizon \\
DreamerV3 (official) & decoded obs & $\|\hat y{-}y^*\|{<}\theta_k$ & speed, horizon \\
V-JEPA 2-AC (official) & token embed. & $\|\hat z{-}z^*\|{<}\theta_k$ & action mag. \\
\;\;$+$ real frames & token embed. & $k{=}1$ & class order \\
\bottomrule
\end{tabular}}
\end{table}

\subsection{Online deployment and refusal on hardware}\label{sec:real}

The Franka experiments test online deployment and the transition from reliance to refusal. At $R=9$\,cm, the complete settlement--credit--audit pipeline runs online and supports replay of all 1{,}062 prediction uses (Appendix~C). At $R=4$\,cm, a prospective study tests how new execution failures change reliance under a fixed threshold. Task success and wall time measure the consequences of the deployed fallback.

\emph{Prospective protocol.}
The robot pushes four object classes using a wrist RGB-D camera and AprilTag-based endpoint settlement. The mechanism study uses the retrained vision model from the deployment study, prediction tolerance $R=4$\,cm, and a threshold fixed at $\tau=0.816$ from the earlier measured cost anchor $1-4.0/21.7$. Task success is entry into a 6-cm goal disk within 14 actions. Three arms follow one block-randomized schedule. Each maintains its own persistent history, initialized from the same warmup log re-settled at 4\,cm. Ledger thresholds the empirical agreement rate directly; Blind permits every candidate; Random refuses with a fixed probability of 0.48 on irregular/slippery objects and zero otherwise. A refused candidate is replaced by a probe with its own registered endpoint prediction.

The completed study comprises 77 episodes: 63 main episodes (21 per arm) and 14 degradation episodes. Task comparisons use the main sample, while the degradation episodes are reported separately. Appendix~D gives the collection rule, protocol changes, and analysis details.

\begin{figure}[t]
\centering
\includegraphics[width=\linewidth]{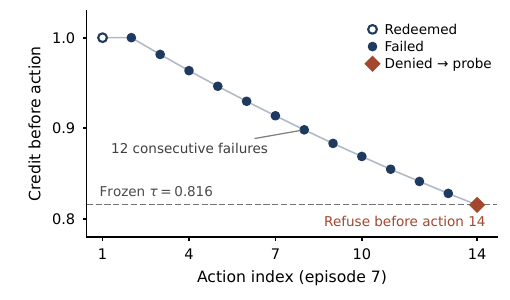}
\caption{Online refusal in main episode 7 (Ledger, cylinder). The first prediction matches the observed endpoint; the next 12 consumed predictions fail at 5.45--6.27\,cm. Before action 14, credit reaches $53/65=0.8154<\tau=0.816$, and the gate denies the candidate and invokes a probe. Each estimate uses only preceding settlements, linking the credit trajectory to the pre-action gate decision.}
\label{fig:hwep7}
\end{figure}

\emph{A previously high-credit class becomes refused.}
Episode 7 shows how accumulated support determines the intervention time. Cylinder credit begins at 1.00, backed by 52 agreements during warmup. One additional agreement and twelve consecutive failures bring the estimate to $53/65=0.8154$, below the frozen threshold, so the gate refuses the candidate before action 14 (Fig.~\ref{fig:hwep7}). The predicate and threshold remain fixed, and each estimate uses only preceding settlements. The twelve failures are therefore the new evidence required to change the decision given the cell's supporting history. Refusal occurs with one action left in the episode budget, connecting the timing of intervention to the evidence accumulated before commitment. Appendix~D records the final probe and episode outcome.

\begin{table}[t]
\centering
\caption{Prospective hardware follow-up: 63 main episodes. Success intervals are episode-level Wilson 95\% CIs; other entries are means per episode. Post-repair episodes (28--62) are a subset of the full sample. Burns exclude probe prediction failures.}
\label{tab:hwfollowup}
\footnotesize
\setlength{\tabcolsep}{2.2pt}
\resizebox{\linewidth}{!}{%
\begin{tabular}{llrcrrr}
\toprule
Sample & Arm & $n$ & Success & Burns & Probes & Wall (s) \\
\midrule
\multirow{3}{*}{All main} & Ledger & 21 & 0.86 [0.65,0.95] & 1.19 & 3.24 & 163 \\
 & Blind & 21 & 0.95 [0.77,0.99] & 1.19 & 0.00 & 117 \\
 & Random & 21 & 1.00 [0.85,1.00] & 0.86 & 0.71 & 105 \\
\midrule
\multirow{3}{*}{Post-repair} & Ledger & 11 & 1.00 [0.74,1.00] & 0.73 & 2.73 & 136 \\
 & Blind & 12 & 0.92 [0.65,0.99] & 1.58 & 0.00 & 130 \\
 & Random & 12 & 1.00 [0.76,1.00] & 0.42 & 0.58 & 94 \\
\bottomrule
\end{tabular}}
\end{table}

\emph{Task outcomes and probe behavior.}
The task consequences of refusal also depend on the action that follows it. Across the main sample, Ledger completes 18/21 tasks, compared with 20/21 for Blind and 21/21 for Random (Table~\ref{tab:hwfollowup}). Ledger and Blind both incur 1.19 burns/episode; their paired wall-time difference is $+46$\,s (95\% interval $[-7,104]$). Against Random, Ledger takes $+58$\,s ($[5,108]$). The pre-specified task comparisons retain every main episode, spanning the probe adjustment described below.

The initial 1.5-cm probe could stop short of the slippery object because the per-class radial profile overestimated its facing extent. Penetration was increased to 5\,cm from episode 28 as a documented protocol change; all earlier trials remain in the full sample. Before the change, Ledger succeeds in 7/10 episodes; afterwards it succeeds in 11/11, compared with 11/12 for Blind and 12/12 for Random. In the post-repair paired Ledger--Blind comparison, the burns difference is $-0.82$/episode ($[-2.18,0.36]$) and the wall-time difference is $+6$\,s ($[-45,52]$). The post-repair subset diagnoses fallback behavior across a temporal change within one deployment. With 21 episodes per arm and 11 post-repair pairs, task-success differences remain imprecisely estimated.

After repair, Ledger incurs 0.73 burns/episode versus 1.58 for Blind. Counting probe prediction failures as well gives 1.64 versus 1.58 failures/episode, a post-hoc diagnostic. Burned imagination measures exposure through predictions consumed by the controller, while the inclusive count also captures errors during verification.

\emph{Refusal placement.}
In the main sample, 61/68 Ledger denials occur on the initially low-credit irregular/slippery classes; after repair the count is 29/30. These descriptive counts reflect both initial settlement support and subsequent settlements. Random's fixed refusal probability yields 0.71 denials/episode, compared with 3.24 for Ledger. The hardware counts describe the deployed policies, while matched-refusal simulation comparisons isolate history's contribution to intervention placement (Sec.~\ref{sec:gate}). Probe settlements evaluate the predictions registered for verification actions and do not measure the counterfactual outcomes of the original denied candidates, which remain unexecuted. The episode-7 trace supplies the complementary temporal evidence: new settlements change the gate decision before the next action.

\subsection{Design checks and supporting analyses}\label{sec:supporting}

\textbf{Drift and conservatism.}
Across a seven-round drift series spanning scene shifts, same-sensor miscalibration, and model staleness, the violation rate of consumed claims remains nearly unchanged under the conservative gate. This remains true after replacing the strong model with the weak model (0.218 vs.\ 0.221). Because low-credit claims are withheld from consumption, degradation instead appears in task metrics and burned imagination. Drift detection is therefore more informative on an unshielded stream. This comparison distinguishes predictor drift from reference drift: diagnosing a change in the settlement sensor itself requires an independent reference.

\textbf{Externalization and audit.}
In an engineered occluded pocket behind a low-texture doorway, a sparsely supported cell produces 45 denials and no prediction use, triggering three observation actions before commitment. This experiment provides the clearest controlled example in which denial directly causes information acquisition. Simulation also validates dependency-record completeness and tamper detection through 165/165 independent log replays and settlement-hash mismatch detection.

\textbf{Feedback after refusal.}
A log-level audit tests the proposed refusal$\rightarrow$observation$\rightarrow$settlement feedback path. In healthy 2D navigation at $\tau=0.6$, 64\% of denied ledger claims settle within the same episode and 40\% trigger an observation within five steps; among the observed subset, denied claims fail at 0.326 versus 0.208 for allowed claims, with the same ordering in 10/10 seeds. The book-blind verifier settles 43\% of its denials, triggers an observation for only 8\%, and does not separate their failure rates (0.210 vs.\ 0.206). The same audit measures observation capacity under localized degradation: 4--5\% of ledger denials settle in-episode when refusal volume exceeds the observation budget. These post-hoc results quantify the feedback acquired after refusal. Denied-claim failure rates are conditional on the subset that becomes observable.

\textbf{Task value and cost sensitivity.}
Burn rate measures prediction exposure per consumed claim. Task value additionally depends on verification, completion, and collisions. A post-hoc analysis puts these outcomes on a common scale,
\[
\begin{aligned}
J={}&\mathrm{burns}+\rho_{\mathrm{look}}\mathrm{looks}
+\rho_{\mathrm{fail}}(1-\mathrm{success})\\
&+\rho_{\mathrm{coll}}\mathrm{collisions},
\end{aligned}
\]
where each $\rho$ expresses a cost relative to one burn. The earlier hardware timing anchor gives $\rho_{\mathrm{look}}=4.0/21.7=0.184$; the comparisons below set $\rho_{\mathrm{coll}}=0$.

Against the book-blind and episode-local gates in healthy navigation, the ledger is cheaper with 95\% CI support through the evaluated $\rho_{\mathrm{fail}}=50$. At $\rho_{\mathrm{fail}}=100$, the point estimates still favor it but both intervals include zero. Point estimates break even near 110 and 133 burns, respectively, beyond which the other gates' higher reach dominates. In manipulation at protocol-selected thresholds, the ledger is cheaper than book-blind at the evaluated $\rho_{\mathrm{fail}}=1,2$. At 5, the difference is $+.256$ with a borderline interval $[.0004,.521]$; from 10 onward, intervals include zero throughout the tested grid. The point-estimate break-even is 14.3 burns. At equal fixed thresholds, the point-estimate crossover is approximately 3.5, illustrating the additional dependence on operating-point selection.

The random-refusal comparison has a different cost profile. In navigation, the ledger becomes cheaper once task failure costs roughly 25 burns. In pooled manipulation, random refusal incurs $+.102$ burns/episode (seed-level 95\% interval $[.022,.182]$), $+.276$ probes/episode ($[.206,.345]$), and $+.038$ task-failure probability ($[.010,.067]$). Nonnegative weighting of these measured components favors the ledger over random refusal. In the paired quadrotor study, task-outcome intervals leave cost ordering unresolved outside the low-$\rho_{\mathrm{fail}}$ region.

These regions express conditional policy choices. Task-failure cost was not measured, and the later hardware probes have different timings from the earlier observation anchor. The analysis therefore supplies neither a universal task-value ranking nor a measured hardware cost advantage. It specifies which completion and verification costs would make the observed refusal-placement trade-off useful.

\textbf{Runtime overhead.}
On one CPU core, credit estimation, gating, and registration require 40\,$\mu$s per consumed claim, while settlement requires 823\,$\mu$s, including online head updates and amortized calibrator refits. The complete claim lifecycle therefore takes less than 1\,ms. Candidate search uses raw model outputs, so credit estimation scales with the number of claims in the committed plan rather than the number of candidates.

\textbf{Granularity, partition, and threshold sensitivity.}
Three offline analyses of the logged quadrotor settlement streams examine the ledger's design degrees of freedom (Table~\ref{tab:granularity}); all estimates are prequential and use episode-block bootstrap CIs. First, increasing granularity from a single global bin to the deployed condition$\times$region$\times$horizon partition monotonically improves risk--coverage discrimination (weak stream AURC $0.310\rightarrow0.221$; strong $0.268\rightarrow0.229$; medium $0.277\rightarrow0.232$), while per-bin calibration degrades (weak ECE $0.025\rightarrow0.081$). With a fixed amount of data, finer partitions therefore improve discrimination at the cost of thin-bin calibration. The Mondrian merge ladder and fusion head for sparse cells (Table~\ref{tab:sources}) are designed to address this gap. Second, wrong-partition controls test whether discrimination follows meaningful context or simply dividing samples into bins. A fully irrelevant partition returns discrimination to the global level and has the worst calibration of all configurations. Corrupting only the condition axis degrades more gradually, retaining the discrimination supplied by the remaining valid axes. The resulting comparisons measure the contribution of each contextual axis alongside its calibration cost. 

Third, threshold sensitivity depends on the score being thresholded. For the calibrated credit used by the gate (the warmup stream in Fig.~\ref{fig:rc}), failure among retained predictions meets the pre-stated fail$\leq20\%$ target for $\tau\in[0.74,0.77]$ (coverage $.32$--$.20$). For the bins-only full-partition score used in the preceding granularity analyses, the target is met from $\tau=0.70$ onward (failure $.168$--$.180$ over $[0.70,0.80]$). Given the selection sample size (147 retained predictions at $\tau=0.75$; Wilson 95\% interval $[.12,.25]$), the boundaries of this range cannot be statistically resolved. The frozen operating point is therefore validated through closed-loop outcomes across seeds (Table~\ref{tab:headline}), with the warmup interval describing its selection uncertainty.

\begin{table}[t]
\centering
\caption{Partition sensitivity on the weak quadrotor settlement stream (540 settlements; prequential estimates with episode-block bootstrap 95\% CIs). Valid and irrelevant partitions test contextual discrimination; ECE tracks the estimation cost of dividing a fixed sample among finer cells.}
\label{tab:granularity}
\footnotesize
\setlength{\tabcolsep}{4pt}
\begin{tabular}{lccc}
\toprule
Partition & Bins & AURC$\downarrow$ & ECE$\downarrow$ \\
\midrule
Global & 1 & .310 [.25,.38] & .025 \\
Condition & 3 & .262 [.21,.32] & .041 \\
Condition$\times$horizon & 12 & .255 [.20,.31] & .055 \\
Full (cond.$\times$region$\times$horizon) & 39 & \textbf{.221} [.18,.27] & .081 \\
\midrule
Pseudo-cond.$\times$region$\times$horizon & -- & .230 [.19,.28] & .113 \\
Fully random partition & 39 & .251 [.19,.32] & .115 \\
\bottomrule
\end{tabular}
\end{table}

\section{Limitations and Conclusion}\label{sec:limits}

\textbf{Limitations.}
The execution history estimates deployment-specific claim reliability. Transferring on-policy estimates to unexecuted candidates requires within-cell exchangeability; the oracle audit quantifies this transfer gap, while randomized execution directly measures denied-candidate risk on the audited decision stream (Secs.~\ref{sec:dose} and~\ref{sec:auditshift}). Observation budgets determine how much new evidence low-credit cells acquire, and inverse-probability weighting uses attributable settlements to estimate candidate risk within that stream. The post-warmup shifts test adaptation to newly emerging failures; recovery after conditions improve and auditing schedules that balance information against execution cost remain directions for further study. Attribution also depends on state-estimation quality, with discard rates recording outcomes that remain unattributable.

The evaluation covers a compact 2D domain, simulated aerial navigation, and pushing manipulation. Other actions, such as grasping, require corresponding typed settlement predicates. Predicate and granularity specify the quantity whose reliability the ledger estimates; task completion is evaluated as a separate outcome. The hardware deployment study establishes online execution and audit, and the prospective study establishes credit-driven refusal under a frozen threshold. Their task comparisons are conditional on one evolving deployment and its available fallback. With cumulative histories, prior support determines how much new evidence is needed to change a decision; physical probe effectiveness determines what follows refusal. Repeated deployments and alternative fallback policies would test the resulting task trade-offs more broadly. Foundation-model settlement on real frames currently covers offline, one-step ($k=1$) evaluation.

Declared contexts determine the resolution at which reliability is estimated; settlement support determines whether that resolution can be used. An unsupported context receives history features from the prior $(p_0,N,L,U)=(0.5,0,0,1)$ and undergoes the ordinary credit test. Over the 60-episode quadrotor threshold-selection stream, only 6--7 of the 39 condition$\times$region$\times$horizon cells reach the $n{=}25$ merge threshold, with a median of 4--5 settlements per cell. The 15-episode named-gate warmup is sparser still: each arm settles only 118--175 claims, and roughly three quarters of estimates originate in exact cells with fewer than 25 settlements. Most estimates therefore pool evidence across regions or conditions. Region-level persistence is evaluated in the 2D domain with dense histories (Tables~\ref{tab:persist} and~\ref{tab:matched}).

A coarser quadrotor partition with 40 collection episodes raises the median to 16.5--18.5 settlements per cell, still below the merge threshold. At protocol-selected operating points, the ledger incurs fewer burns than episode-local books in 7 of 8 tested cells. Comparisons at this density remain sensitive to seed and operating point. The quadrotor comparisons therefore characterize pooling and sample sensitivity, while the experiments with dense histories supply the placement evidence. Future work includes recurrent manipulation, credit-directed model repair, personalization of gated reliance policies to users and tasks~\cite{li2026personalize}, and studies of credit externalization and operator trust. Code, logs, and deployment records will be released.

\textbf{Conclusion.}
\sys{} connects past execution to future reliance through a persistent, auditable credit record. Matched-refusal controls show how that history changes refusal placement; randomized audits measure the risk of denied candidates, and post-warmup shifts identify the contribution of newly accumulated evidence. The independent manipulation replication supports lower burn rate and probe use than random refusal. On a Franka, complete replay validates online deployment, and a prospective trace shows new failures triggering refusal under a frozen threshold. Task value depends on the comparator, fallback, and cost of an unfinished task; execution history informs which predictions the robot consumes and when it stops relying on them.

\section*{Appendix}

\subsection*{A. Statistical comparison protocol}
All closed-loop quadrotor experiments run in Isaac Sim with RTX-rendered depth. As in most GPU-rendered simulation stacks, rendering varies across runs; episode outcomes are therefore samples from a per-world outcome distribution rather than bitwise-replayable trajectories. We consequently formulate every behavioral comparison as a paired statistical contrast rather than a trajectory comparison. Compared arms share world seeds episode-by-episode, and all arms in a table follow the same protocol and warmup structure. Whenever we claim a difference, per-episode paired deltas enter a hierarchical bootstrap that resamples seeds and then paired episodes within each seed; $10^4$ resamples produce the reported 95\% intervals. Ledger estimates are prequential throughout. Every online credit estimate uses only evidence available before the corresponding outcome. The artifact released with the paper will include audit logs, per-episode records, settlement streams, and analysis scripts, allowing each reported value to be traced to its deployment record.

The persistence battery in Sec.~\ref{sec:gate} was pre-registered: arm grids, endpoints, and pass criteria were frozen before its data were collected. The record dates one pre-run amendment and two post-run deviations, preserving the timing of each analytic choice. The 610-run main battery uses a fixed CPU-cluster environment and is not mixed with the locally collected snapshot in Table~\ref{tab:persist}; the 90-run warmup-density ladder is reported separately. The pre-registered endpoint was absolute burns. Burn rate (burns per consumed prediction) was added post hoc after observing the confounding effect of consumption volume, and both conventions are reported. For the matched frontier, pooled bootstrap intervals are cross-checked using seed-aggregated $t$-intervals because thresholds within a seed are dependent.

The AURC replication, density-stratified AURC, Beta warmup ladder, refusal-feedback audit, unified cost map, randomized-auditing study, and post-warmup-shift study were added post hoc. Each new audit or shift design froze its arms, grid, endpoints, and interpretation rule before its own runs; the $+15\%$ shift extends the initial severe-shift study, and both levels are reported. The AURC studies test whether an original sample ordering replicates; the passive feedback audit characterizes observed denials; and the cost map evaluates dependence on an unmeasured task-failure cost. The targeted manipulation replication likewise froze its independent seeds, expanded random-$p$ grid, endpoints, and interpretation rule before collection. Its original, coverage-completed, independent, and pooled blocks are reported separately. The hardware follow-up is a separate prospective study motivated by the offline tolerance ladder, with dated protocol changes retained in Appendix~D.

Randomized-audit intervals resample paired seed-level failure rates. The inverse-probability-weighted estimate is $(F_a+F_d/\varepsilon)/(N_a+N_d/\varepsilon)$, where $F$ and $N$ count failed and attributable settled claims, respectively, and subscripts $a$ and $d$ denote allowed and audited candidates. It targets the settlement-supported candidate stream encountered by each auditing policy. Shift intervals resample paired seed--threshold contrasts over the three-point grid; seed-aggregated checks also support the severe-shift burn-rate reduction and the $+15\%$ shift's absolute-burn reduction. Time-course values are per-episode means within the first and last ten gated episodes. The shift criterion requires a warm-only minus online burn-rate interval above zero at two or more thresholds; the severe shift meets it at all three, while the $+15\%$ extension has a resolved interval at $\tau=0.6$.

\subsection*{B. Partition sensitivity at scale: 2D-navigation replication}
The granularity and wrong-partition analysis in Table~\ref{tab:granularity} uses roughly 540 settlements per arm. The 2D-navigation domain provides a higher-powered replication with roughly 11{,}000 settlements per arm (Table~\ref{tab:nav2dgran}; episode-block bootstrap CIs are approximately $\pm0.02$). Both findings replicate with non-overlapping intervals. Discrimination improves monotonically with valid granularity (strong AURC $0.202\rightarrow0.129$), while a fully random partition returns to the global level (0.201 vs.\ 0.202) and has the worst calibration. Corrupting only the condition axis reduces discrimination ($0.129\rightarrow0.181$). The higher-volume replication additionally evaluates calibration across granularities: with this settlement support, ECE remains between $0.007$ and $0.03$ at every granularity. The granularity--calibration trade-off in Table~\ref{tab:granularity} is therefore a consequence of sparse histories that diminishes as histories become dense. The Mondrian merge ladder and fusion head bridge this transition.

\begin{table}[t]
\centering
\caption{2D-navigation partition sensitivity (AURC$\downarrow$; $\approx$11k settlements per arm; prequential estimates). Valid granularity improves performance monotonically, whereas irrelevant partitions revert to the global level.}
\label{tab:nav2dgran}
\footnotesize
\setlength{\tabcolsep}{5pt}
\begin{tabular}{lccc}
\toprule
Partition & Strong & Medium & Weak \\
\midrule
Global & .202 & .163 & .333 \\
Condition & .137 & .086 & .307 \\
Condition$\times$horizon & .131 & .085 & .302 \\
Full (cond.$\times$region$\times$horizon) & \textbf{.129} & \textbf{.080} & \textbf{.296} \\
\midrule
Pseudo-cond.$\times$region$\times$horizon & .181 & .137 & .317 \\
Fully random partition & .201 & .184 & .343 \\
\bottomrule
\end{tabular}
\end{table}

\subsection*{C. Hardware deployment and replay analysis}

The deployment study uses four object classes (two physical instances per class), a wrist-mounted RGB-D camera, AprilTag settlement, confidence gating, and a per-frame homography. Its frozen endpoint-region predicate is $R=\max(9\,\mathrm{cm},2.5\times\mathrm{noise\ floor})=9$\,cm; measured settlement noise is 0.7\,mm median (90th percentile 2.0\,mm). Warmup risk--coverage selects frozen thresholds of 0.497/0.499 for the simulation-trained/retrained models. The retrained model uses the same architecture fitted to 300 real pushes (256 training, 44 validation); the two models' shared-validation errors are 2.83 and 1.74\,cm.

Across seven warmup, gated, blind, and scripted-segment logs, replay reconstructs all 1{,}062 records of prediction use: 949 attributable settlements, 108 discards, and five claims pending at episode termination. At 9\,cm, warmup failure rates are 0.000 and 0.004. The logs jointly exercise online registration, attribution, settlement, credit updates, and decision replay. A high-credit episode uses nine predictions without additional observations; an occluded-target episode establishes no usable claim and acquires five observations before acting. The gated and blind logs were collected sequentially with different placements and serve as deployment traces; Sec.~\ref{sec:real} uses a block-randomized schedule for the prospective mechanism study.

\emph{Offline tolerance ladder.}
Stored continuous endpoint errors permit re-settlement at tighter tolerances (Table~\ref{tab:ladder}). For the retrained model, the cost anchor denies 20 consumed predictions at 5\,cm, all on slippery objects, and 70 at 4\,cm, all on irregular/slippery objects. The denied predictions fail more often than the allowed predictions in this replay: 0.200 vs.\ 0.119 at 5\,cm and 0.257 vs.\ 0.102 at 4\,cm. These class-level estimates and outcomes come from the same deployment pool. The resulting offline tolerance diagnosis informed the prospective mechanism study in Sec.~\ref{sec:real}, where closed-loop outcomes are measured.

\begin{table}[t]
\centering
\caption{Offline re-settlement of hardware logs at tighter claim tolerances $R$. Entries are class-level estimates from the same replayed deployment pool (retrained model, 146 gated consumed predictions; simulation-trained model in parentheses). For the retrained arm, the final columns report realized would-fail rates among consumed predictions that $\tau_{\mathrm{dt}}$ denies or allows. The table describes offline credit re-estimation of the observed endpoint errors; Sec.~\ref{sec:real} reports the prospective closed-loop evaluation.}
\label{tab:ladder}
\footnotesize
\setlength{\tabcolsep}{3pt}
\resizebox{\linewidth}{!}{%
\begin{tabular}{lccrrcc}
\toprule
$R$ & Agreement & Lowest class & \multicolumn{2}{c}{Denials of 146 (75)} & \multicolumn{2}{c}{Would-fail rate} \\
 & retr.\ (sim) & retr.\ (sim) & $\tau{=}0.50$ & $\tau_{\mathrm{dt}}{=}0.82$ & denied & allowed \\
\midrule
9\,cm & .996 (.970) & boxy .98 (slip.\ .90) & 0 (0) & 0 (0) & -- & .000 \\
7\,cm & .988 (.963) & boxy .95 (slip.\ .90) & 0 (0) & 0 (0) & -- & .008 \\
5\,cm & .942 (.872) & slip.\ .85 (irreg.\ .78) & 0 (0) & 20 (0) & .200 & .119 \\
4\,cm & .843 (.799) & slip.\ .70 (irreg.\ .70) & 0 (0) & 70 (20) & .257 & .102 \\
3\,cm & .735 (.645) & slip.\ .57 (irreg.\ .56) & 19 (0) & 114 (75) & .299 & .000 \\
\bottomrule
\end{tabular}}
\end{table}

\begin{figure}[t]
\centering
\includegraphics[width=0.92\linewidth]{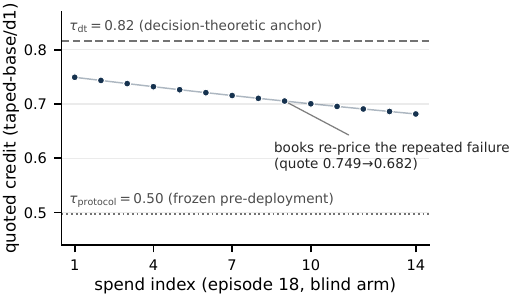}
\caption{Deployment trace: online credit revision under repeated endpoint error. Fourteen identical no-contact pushes in a blind-arm episode leave the object stationary, and each claim fails at $\approx9.5$\,cm. Successive settlements lower the consumed cell's credit; the lines mark the deployment threshold and the later cost anchor. The blind-arm trace records estimator updates during continued consumption.}
\label{fig:ep18}
\end{figure}

\emph{Credit revision during repeated execution.}
In blind episode 18, fourteen repeated no-contact pushes each fail at approximately 9.5\,cm (Fig.~\ref{fig:ep18}). Calibrated credit decreases from 0.749 to 0.682, with each estimate backed by 83--96 prior settlements in the consumed cell, including 67 from warmup. The blind-arm trace isolates online estimator updates during repeated consumption. Figure~\ref{fig:ep18} locates these estimates relative to the deployment threshold $\tau=0.50$ and the post-hoc cost anchor $\tau_{\rm dt}=1-C_{\rm look}/C_{\rm burn}=0.82$. The measured costs are $C_{\rm burn}=21.7$\,s and $C_{\rm look}=4.0$\,s, the latter dominated by a declared observation dwell. Offline re-settlement at 4\,cm makes the empirical agreement rate cross that anchor on the thirteenth failure; at 3\,cm it starts below the anchor. This separates the effects of the settlement predicate, credit estimator, and intervention threshold.

\subsection*{D. Prospective hardware protocol}

The follow-up fixes the model, $R=4$\,cm, $\tau=0.816$, empirical agreement rate, and three-arm schedule before the main trials. The pre-specified empirical estimator uses the archived warmup endpoint errors directly to initialize the 4-cm histories. Re-settling only that warmup log yields boxy $53/57$, cylinder $52/52$, irregular $62/84$, and slippery $44/63$ agreements. Every arm starts from these same histories and updates them from its own attributable settlements, including probes; on resumption, its prior settlements restore the history.

The completed sample contains 63 main episodes in blocks M00--M06 and 14 degradation episodes, under the pre-declared one-day fallback. Main arms share scheduled instances and goal coordinates within blocks. The zero-action episode 1 is retained; a start-distance rule of at least 12\,cm was added after episode 2. Degradation goals were generated relative to the placed object after a documented protocol change, and this block is reported separately: all 14 finalized episodes succeeded (Ledger 6/6, Blind 4/4, Random 4/4). Six pilot episodes and six later video-demonstration episodes are excluded from the statistical sample.

Probe penetration was increased from 1.5 to 5\,cm after episode 27, a dated deviation on September 6. The full sample retains the earlier outcomes: Ledger episodes 11 and 19 each exhausted all 14 actions on probes without completing the task. In episode 7, the gate refused before action 14; the final probe also failed settlement, and the episode ended without reaching the goal. The main analysis reports the complete sample and the subsets before and after the penetration change.

Task success and episode wall time are the frozen co-primary endpoints; burns, probes, refusal placement, and subsequent prediction failures are secondary. Probe predictions and original denied candidates have distinct action conditions; only the executed probe yields an attributable outcome. The inclusive prediction-failure count in Sec.~\ref{sec:real} is a post-hoc diagnostic. Success intervals are Wilson intervals over episodes. Paired differences use 10,000 percentile bootstrap resamples of main episodes matched by block and instance. There are 21 full-sample pairs and 11 post-repair pairs per comparator; unmatched episodes remain in the descriptive means. The intervals are conditional on this single deployment and do not capture all temporal dependence from evolving histories or variation across deployments.

The threshold adopts the original deployment's cost anchor as a fixed operating point. Its 4-s observation term does not measure the follow-up probe: after repair, Ledger's logged action-and-settlement duration averages 27.4\,s for probes and 26.7\,s for ordinary pushes. A probe can also share an ordinary push's penetration setting. The anchor specifies the evaluated operating point; physical probe effectiveness and measured episode cost determine its task consequences.

\bibliographystyle{IEEEtran}
{\footnotesize
\bibliography{refs}}

\end{document}